\def\year{2022}\relax
\documentclass[letterpaper]{article} 
\usepackage{aaai22}  
\usepackage{times}  
\usepackage{helvet}  
\usepackage{courier}  
\usepackage[hyphens]{url}  
\usepackage{graphicx} 
\usepackage{natbib}  
\usepackage{caption} 
\DeclareCaptionStyle{ruled}{labelfont=normalfont,labelsep=colon,strut=off} 
\usepackage{algorithm}
\usepackage{algorithmic}
\usepackage{booktabs}
\usepackage{subfig}
\usepackage{xcolor}
\usepackage{tcolorbox}
\usepackage{inconsolata}

\usepackage{newfloat}
\usepackage{listings}
\floatstyle{ruled}
\newfloat{listing}{tb}{lst}{}
\floatname{listing}{Listing}
\title{Overcoming Language Disparity in Online Content Classification with Multimodal Learning}
\author{
    Gaurav Verma, Rohit Mujumdar, Zijie J. Wang, Munmun De Choudhury, Srijan Kumar
}
\affiliations{
    Georgia Institute of Technology\\
    \{gverma, rohitmujumdar, jayw, munmund, srijan\}@gatech.edu
}

\usepackage{bibentry}

\begin{document}

\maketitle

\begin{abstract}

Advances in Natural Language Processing (NLP) have revolutionized the way researchers and practitioners address crucial societal problems. Large language models are now the standard to develop state-of-the-art solutions for text detection and classification tasks. However, the development of advanced computational techniques and resources is disproportionately focused on the English language, sidelining a majority of the languages spoken globally. While existing research has developed better multilingual and monolingual language models to bridge this language disparity between English and non-English languages, we explore the promise of incorporating the information contained in images via multimodal machine learning. Our comparative analyses on three detection tasks focusing on crisis information, fake news, and emotion recognition, as well as five high-resource non-English languages, demonstrate that: \textit{(a)} detection frameworks based on pre-trained large language models like BERT and multilingual-BERT systematically perform better on the English language  compared against non-English languages, and \textit{(b)} including images via multimodal learning bridges this performance gap. We situate our findings with respect to existing work on the pitfalls of large language models, and discuss their theoretical and practical implications. 

\end{abstract}

\section{Introduction}
Users of social computing platforms use different languages to express themselves~\cite{mocanu2013twitter}. These expressions often give us a peek into personal-level and societal-level discourses, ideologies, emotions, and events~\cite{kern2016gaining}. It is crucial to model \textit{all} of these different languages to design equitable social computing systems and to develop insights that are applicable to a wider segment of the global population.

\begin{figure}
    \centering
    \includegraphics[ width=0.47\textwidth]{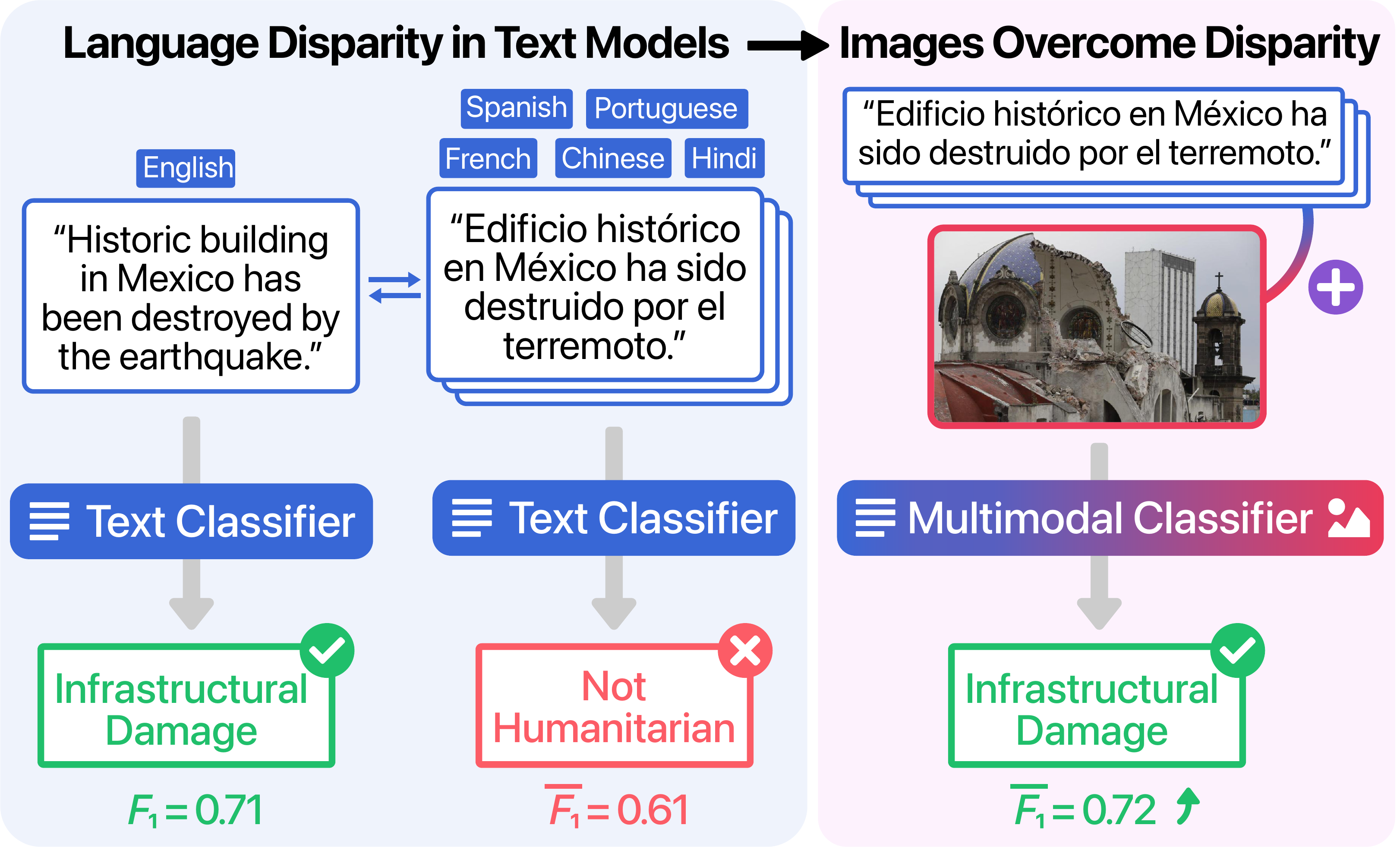}
    \caption{{Overview figure.} We use multimodal (image + text) learning to overcome the language disparity that exists between English and non-English languages. The figure illustrates an example of a social media post that is correctly classified in English but misclassified in Spanish. Including the corresponding image leads to correct classification in Spanish as well as other non-English languages. }
    \label{fig:overview}
\end{figure}

In recent years, we have seen remarkable ability in using linguistic signals and linguistic constructs extracted from social media and web activity toward tackling societal challenges, whether in detecting crisis-related information~\cite{houston2015social} or identifying depression-related symptoms
~\cite{de2013predicting}. 
While earlier approaches relied on qualitative language inference techniques~\cite{crook2016content}, using pre-existing dictionaries~\cite{pennebaker2001linguistic}, and traditional classifiers ~\cite{glasgow2014our},  more recent approaches leverage the advances in deep learning-based language modeling techniques. Large pre-trained models like BERT~\cite{DBLP:journals/corr/abs-1810-04805} are frequently used to train classifiers in tasks pertaining to social good~\cite{singhal2019spotfake, sun2019utilizing} and are now a new standard to build state-of-the-art classification systems to support real-world decision-making.

As \citeauthor{joshi2020state} (\citeyear{joshi2020state}) illustrate, these rapidly evolving language technologies and their applications are largely focused on only a very small number of over 7000 languages in the world. A majority of the research in natural language processing (NLP) is focused on a few high resource languages, and disproportionately on English~\cite{sabrina2016language, bender2019rule}. The development of systems that can model languages beyond English is important for ensuring \textit{(a)} inclusion of communities, \textit{(b)} equitable extension of services that are driven by these language technologies to diverse groups, and \textit{(c)} preservation of endangered languages~\cite{muller2021being}. Especially in the context of social computing, language-specific lapses can lead to inequitable outcomes. For instance, lower detection abilities on Twitter posts published in Spanish could possibly lead to inequitable humanitarian interventions in times of crisis; and, the lack of powerful misinformation detectors for the Chinese language can possibly lead to situations where specific-language speaking individuals are more vulnerable to health-related misinformation.
As BERT-like monolingual and multilingual models take a central role in building approaches to address crucial societal tasks, the bias toward the English language can propagate, reinforce, and even exacerbate the existing inequities that many underserved groups face~\cite{pew_latinos}.

Existing attempts to bridge this gap between English and non-English languages have focused on developing better multilingual and monolingual (non-English) language models~\cite{nozza2020mask}. In this work, we explore the promise of information that lies in other complementary modalities, specifically images~(\ref{fig:overview}). Considering images as an additional modality has proven to be beneficial in a wide range of scenarios --- from accurately estimating dietary intake in a pediatric population~\cite{higgins2009validation}, to creating effective questionnaires~\cite{reynolds2011picture}. The underlying idea stems from the simple fact that images are not bound by any language. We propose the use of multimodal learning, which jointly leverages the information in related images and text, to boost performance on the non-English text and effectively bring it closer to the performance on English text. More concretely, we study the following two research questions in this work:

\vspace{0.05in} \noindent\textbf{RQ1}: \textit{Does using large language models for social computing tasks lead to lower performance on non-English languages when compared to the English language?}

\vspace{0.02in} \noindent\textbf{RQ2}: \textit{Can inclusion of images with multimodal learning help in bridging the performance gap between English and non-English models? }

{To this end, we study the performance of fine-tuned BERT-based monolingual models and multilingual-BERT on three distinct \textit{classification tasks} that are relevant to social computing:} \textit{(i)} humanitarian information detection during crisis~\cite{ofli2020analysis}, \textit{(ii)} fake news detection~\cite{shu2017fake}, and \textit{(iii)} emotion detection~\cite{duong2017multimodal}. These tasks involve categorizing posts/articles published on the web into real-world concepts that help determine, for instance, the type of humanitarian effort required during a crisis or the veracity of published news. Besides English, we consider five high-resource languages: Spanish, French, Portuguese, (Simplified) Chinese, and Hindi. Via extensive comparative analysis on these existing datasets, we demonstrate that {\textit{(a)} large language models --- whether monolingual or multilingual --- systematically perform better on English text compared to other high-resource languages}, and \textit{(b)} incorporating images as an additional modality leads to considerably lesser deviation of performance on non-English languages with respect to that on English\footnote{Project webpage with resources: \url{https://multimodality-language-disparity.github.io/}}. We conclude by discussing the implications of these findings from both practical and theoretical standpoints, and situate them with respect to prior knowledge from the domains of NLP and social computing. 

\section{Related Work}
We discuss three major themes of research that are relevant to our work: the use of large language models in developing approaches for social computing tasks, the discussion of the pitfalls of large language models and their treatment of non-English languages, and the role of multimodal learning in developing social media classification systems. 

\vspace{0.05in} \noindent\textbf{Large language models for social computing tasks:}
Development and deployment of large language models --- deep learning models trained on massive amounts of data collected from the web, have transformed not only the field of NLP but also related fields that leverage text data to make inferences~\cite{rasmy2021med}. To this end, large language models have been used for various applications in social computing ~\cite{arviv2021sa, choi2021more}. The effectiveness of language models in addressing these tasks can be primarily attributed to two factors: (i) they are trained on massive amounts of unannotated text data, leading to a general understanding of natural language, and (ii) they can be easily fine-tuned for specific tasks with moderately-sized annotated data to demonstrate task-specific understanding. Several language models such as BERT~\cite{DBLP:journals/corr/abs-1810-04805} and T5~\cite{raffel2020exploring} have been developed for the English language. Since these models cover only English, large multilingual variants like mBERT~\cite{DBLP:journals/corr/abs-1810-04805} and mT5~\cite{xue2021mt5} have also been developed to model over a hundred other languages beyond English. These language models (both monolingual and multilingual) are widely adopted to develop state-of-the-art approaches for several tasks where the textual modality withholds key information. 

\begin{figure*}[!th]
    \centering
    \includegraphics[width=\textwidth]{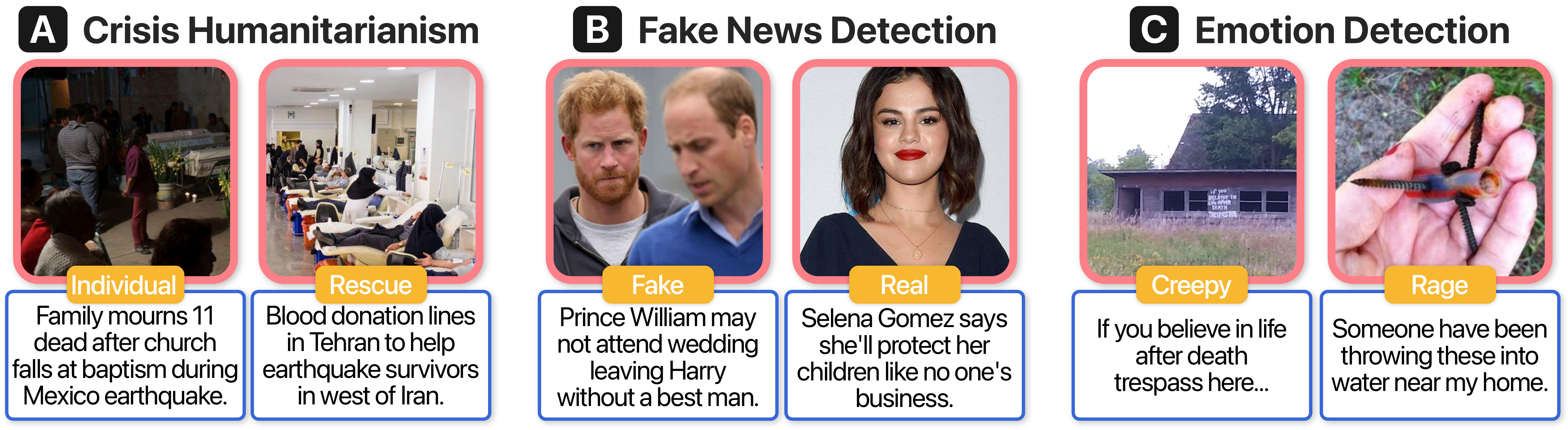}
    \caption{{Illustrative examples from considered multimodal datasets.} We consider three classification datasets for our experiments: {(A)} crisis humanitarianism dataset
    (number of classes: $5$; {infrastructure and utility damage: 10\%}, {rescue volunteering or donation effort:14\%}, {affected individuals: 1\%}, {other relevant information:22\%}, \& {not humanitarian:53\%}), {(B)} fake news detection dataset 
     (number of classes: $2$; {fake:21\%} \& {real:79\%}), and {(C)} emotion detection dataset (number of classes: 4; {creepy:22\%}, {rage:19\%}, {gore:25\%}, \& {happy:34\%}). 
    }
    \label{fig:dataset_examples}
\end{figure*}

\vspace{0.05in} \noindent\textbf{Language disparity in NLP:}
Scholars have discussed the disproportionate focus in NLP research on the English language ~\cite{bender2019rule, joshi2020state, sabrina2016language}. Since approaches to address social computing tasks are increasingly relying on NLP techniques centered around large language models, it is important to understand the possible implications of this disproportionate focus on the state of social computing research. Prior studies have tried to understand the pitfalls of using large language models --- environmental and financial costs~\cite{strubell2019energy}, reliance on data that represents hegemonic viewpoints~\cite{bender2021dangers}, encoding biases against marginalized populations~\cite{basta2019evaluating}. However, our work focuses on comparing English language models with non-English language models in a social computing context. Similar to English, multilingual variants of language models are used to develop the state-of-the-art\footnote{Leaderboard: \url{https://bertlang.unibocconi.it/}} approaches for multiple high-resource non-English languages~\cite{nozza2020mask}. To this end, previous research has focused on understanding how multilingual language models treat various non-English languages relative to each other, especially the contrast between high-resource and low-resource languages ~\cite{pires2019multilingual, wu2020all, nozza2020mask, muller2021being}. In this work, we do not focus on the general pitfalls of large language models or comparisons across non-English languages. Instead, we aim to establish the language disparity between English and non-English languages that is caused due to the adoption of large language models. 

\vspace{0.05in} \noindent\textbf{Multimodal learning:}
Multimodal learning involves relating information from multiple content sources. On the web, the text is often associated with images, especially on social media platforms like Twitter, Instagram, and Facebook. Multimodal learning allows us to combine modality-specific information into a joint representation that captures the real-world concept corresponding to the data~\cite{ngiam2011multimodal}. To this end, inference based on multimodal learning has demonstrated better performance than both text-only and image-only methods, especially in scenarios where access to complementary information can be crucial (e.g., assessing whether a Twitter post (image + text) is about disaster~\cite{ofli2020analysis}, or if a news article (image + title) is fake~\cite{singhal2020spotfake+}, whether the Reddit post conveys rage~\cite{duong2017multimodal}). However, the studies that demonstrate the effectiveness of multimodal learning do so while making comparisons against language-specific text-only methods, without making any comparisons across different languages. In this work, we aim to use multimodal learning, more specifically images, to bridge the gap between English and non-English languages.  

\section{Datasets}
To achieve robust and generalizable findings, we utilize a comparative analytic approach on three different pre-existing datasets that cover issues like humanitarian information processing, fake news detection, and emotion detection. Figure \ref{fig:dataset_examples} presents some examples from the three datasets discussed below as well as the proportion of classes.

\vspace{0.05in} \noindent\textbf{Multimodal crisis humanitarian dataset:} In times of crises, social media often serves as a channel of communication between affected parties and humanitarian organizations that process this information to respond in a timely and effective manner. To aid the development of computational methods that can allow automated processing of such information and, in turn, help humanitarian organizations in gaining real-time situational awareness and planning relief operations, Alam et al.~\cite{alam2018crisismmd} curated the CrisisMMD dataset. This multimodal dataset comprises $7,216$ Twitter posts (images + text) that are categorized into $5$ humanitarian categories. 
The dataset covers 7 crises that occurred in 2017 all over the globe (3 hurricanes, 2 earthquakes, 1 wildfire and floods). 
We formulate the task of humanitarian information detection as a multi-class classification problem, and use the standardized training ($n = 5263$), evaluation ($n = 998$), and test ($n = 955$) sets in our experiments. We maintain the exact same training, validation, and test splits for all the experiments that involve this dataset.

\vspace{0.05in} \noindent\textbf{Multimodal fake news dataset:} Ease of publishing news on online platforms, without fact-checking and editorial rigor, has often led to the widespread circulation of misleading information on the web~\cite{lazer2018science}. Shu et al. (\citeyear{shu2017fake, shu2018fakenewsnet}) curated the FakeNewsNet dataset to promote research on multimodal fake news detection; it comprises full-length news articles (title + body) from two different domains: politics (fake/real labels provided by PolitiFact) and entertainment (fake/real labels provided by GossipCop)  and the corresponding images in the articles. The fake news detection task can therefore be formulated as a binary classification task, where the \texttt{label:0} corresponds to the \texttt{real} class and the \texttt{label:1} corresponds to the \texttt{fake} class. We use the pre-processed version of the dataset provided by ~\citeauthor{singhal2020spotfake+} (\citeyear{singhal2020spotfake+}) and consider only the title of the news article for our experiments while dropping the body of the article. Furthermore, we combine the two domains (entertainment and politics) to create a single dataset and use the same train and test splits like \citeauthor{singhal2020spotfake+} We, however, randomly split the original train set in $90:10$ ratio to create an updated train and validation set. Effectively, our final train, validation, and test sets comprise $9502$, $1055$, and $2687$ news articles, each example containing the title of the news and an image. 

\vspace{0.05in} \noindent\textbf{Multimodal emotion dataset:} Using user-generated content on the web to infer the emotions of individuals is an important problem, with applications ranging from targeted advertising~\cite{teixeira2012emotion} to detecting mental health indicators~\cite{de2013social}. To this end, we collect the dataset introduced by ~\citeauthor{duong2017multimodal} (~\citeyear{duong2017multimodal}) for the task of multimodal emotion detection. The dataset comprises Reddit posts categorized into 4 emotion-related classes, \texttt{creepy}, \texttt{gore}, \texttt{happy}, and \texttt{rage}, where each post contains an image and text. We crawled the images from Reddit using the URLs provided by the authors and randomly split the dataset in a 80:10:10 ratio to obtain the train ($n = 2568$), validation ($n = 321$), and test ($n = 318$) sets. Similar to other datasets, we maintain the exact same splits for all the experiments that involve this dataset to ensure consistent comparisons. 

\begin{table}[!t]
    \centering
        \resizebox{7cm}{!}{
    \begin{tabular}{l  c c  c}
        {\textbf{Language}} & {\textbf{Fluency}} & {\textbf{Meaning}} & {\textbf{Cohen's $\mathbf{\kappa}$}}\\\midrule
        {Spanish (es) } & {4.01} & {4.10} & {0.81}\\
        {French (fr)} & {4.07} & {4.24} & {0.83}\\
        {Portuguese (pt)} & {3.98} & {4.22} & {0.86}\\
        {Chinese (zh)} &  {4.06} & {4.29} & {0.84}\\
        {Hindi (hi)} & {3.91} & {4.12} & {0.82}\\
    \end{tabular}
    }
    \caption{{{Quality assessment of \textit{machine} translation.} Average scores assigned by human annotators on a 5-point Likert scale ($1-5$) for translation quality of generated text, and the agreement scores between annotators for each language for the fluency scores. $N = 200$ examples per language per dataset; $3$ annotators per example.}}
    \label{tab:human_eval_automated}
\end{table}

\vspace{0.05in} \noindent\textbf{{Curating non-English datasets:}} All the three datasets discussed above only have texts (Twitter posts, news articles, and Reddit posts) in English. Given the lack of non-English multimodal datasets, we employ machine translation to convert English text into different target languages. For translation, we use the MarianNMT system, which is an industrial-grade machine translation system that powers Microsoft Translator~\cite{mariannmt}. As target languages, we consider the following five non-English languages: Spanish (\texttt{es}), French (\texttt{fr}), Portuguese (\texttt{pt}), Simplified Chinese (\texttt{zh}), and Hindi (\texttt{hi}). Together, these five languages represent culturally diverse populations -- minority groups in the United States (Hispanics), Asians, and the Global South, and are written in various scripts -- Latin (\texttt{es}, \texttt{fr}, and \texttt{pt}), Hanzi (\texttt{zh}), and Devanagari (\texttt{hi}). It is worth noting that none of these five non-English languages are considered to be low-resource languages~\cite{hedderich2021survey} -- which is a more appropriate designation for languages like Sinhala, the Fijian language, and Swahili. However, since these languages are sufficiently high-resource languages, MarianNMT can produce high-quality translations in these languages from the original English text. 

We use the pre-trained language-specific translation models of MarianNMT, made available via HuggingFace~\cite{wolf2019huggingface}, to translate the text part of each example in the three datasets to the five target language (\texttt{en} $\rightarrow$ \texttt{es}, \texttt{fr}, \texttt{pt}, \texttt{zh}, \texttt{hi}). Before translating, we pre-processed the English text to remove URLs, emoticons, platform-specific tokens (like `RT' for indicating retweets on Twitter), and symbols like @ and \#. We also expanded negatives like \textit{can't} and \textit{won't} to `\textit{can not}' and `\textit{will not}'. Overall, translating the English text to five non-English languages gives us $6$ different versions of each of the three datasets discussed above, where each version differs only in terms of the language of the text. It is worth emphasizing that the train, validation, and test splits remain the same across different languages; this is done to ensure a meaningful comparison of classification models' performance across different languages.

\vspace{0.05in}\noindent\textbf{{Human-translated subset for crisis humanitarianism:}} {Besides the machine-translated text, we also obtain manual translations for a subset of examples from the test set of the Crisis Humanitarianism dataset. 
For Spanish, French, and Portuguese, we recruited workers from Amazon Mechanical Turk (AMT) who were designated as `Masters' and proficient in both English and the target language. For Chinese and Hindi, we obtained annotations from doctoral students fluent in both English and Chinese/Hindi. The recruited participants translated $200$ examples from the test set for each non-English language.
The annotators were shown both the original Twitter post and were instructed to translate the text to the target language while maintaining grammatical coherence and preserving semantic meaning. We use this manually-translated subset of the test set for evaluation purposes alone --- allowing us to observe the validity of observed trends on a cleaner dataset. Next, we assess the quality of machine- and human-translated text.}

\begin{table}[!t]
    \centering
    \resizebox{7cm}{!}{
    \begin{tabular}{l  c c  c}
        {\textbf{Language}} & {\textbf{Fluency}} & {\textbf{Meaning}} & {\textbf{Cohen's $\mathbf{\kappa}$}}\\\midrule
        {Spanish (es)} & {4.21} & {4.33} & {0.85}\\
        {French (fr)} & {4.19} & {4.29} & {0.82}\\
        {Portuguese (pt)} & {4.08} & {4.36} & {0.79}\\
        {Chinese (zh)} & {4.31} & {4.40} & {0.85}\\
        {Hindi (hi)} & {4.39} & {4.45} & {0.87}\\
    \end{tabular}
    }
    \caption{{{Quality assessment of \textit{human} translation for crisis humanitarianism dataset.} Average scores assigned by human annotators on a 5-point Likert scale ($1-5$) for translation quality of generated text, and the agreement scores between annotators for each language for the fluency scores. $N = 200$ examples per language; $3$ annotators/example.}}
    \label{tab:human_eval_manual}
\end{table}

\begin{figure*}[!t]
    \centering
    \includegraphics[width=1.0\linewidth]{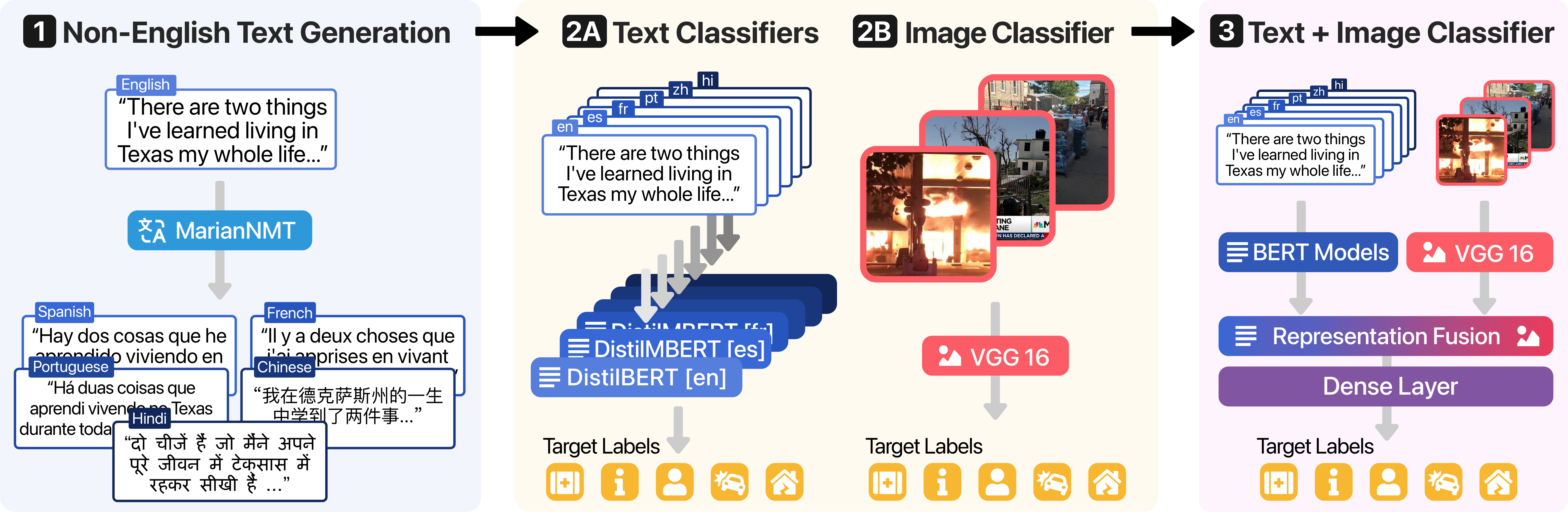}
    \caption{{Overview of the adopted methodology.} After using machine translation to obtain high-quality translations of the English text in our datasets (1), we train language-specific text-only classification models (2A) and  image-only classification models (2B). The multimodal classifier (3) fuses the representations obtained from  trained text-only and image-only models, and predicts the label based on joint modeling of the input modalities.}
    \label{fig:method_overview}
\end{figure*}

\vspace{0.05in} \noindent\textbf{Human evaluation of translation quality:} 
MarianNMT is the engine behind Microsoft Translator, a system that demonstrates translation quality that is close to human parity for specific languages and in constrained settings~\cite{microsoft}. We conduct an independent evaluation of the generated translations of examples from our datasets. {For this, we randomly sampled 200 examples from each dataset (600 examples in total) and asked human annotators to assess the translation quality. Similar to above, the recruited annotators were  AMT workers for Spanish, French, and Portuguese, and doctoral students for Chinese and Hindi.
Each of the 3000 (i.e., 600 $\times$ 5) translation pairs was annotated by $3$ annotators where they responded to the following two questions using a five-point Likert scale (1: strongly disagree, \ldots,  5: strongly agree)}: (i) \textit{Is the $<$Spanish$>$\footnote{The language name was changed as per the target language for which the annotators were rating. {Also, we inserted some ``attention-check'' examples during the annotation process to ensure the annotators read the text carefully before responding. This was done by explicitly asking the annotators to mark a randomly-chosen score on the Likert scale regardless of the original and translated text. We discard the annotations from annotators who did not respond to all the attention-check examples correctly.} }  text a good translation of the English text?}, and (ii) \textit{Does the $<$Spanish$>$ text convey the same meaning as the English text?} While the first question encouraged the annotators (i.e., AMT workers for Spanish, French, and Portuguese, and doctoral students for Chinese and Hindi) to evaluate the quality of the translations, including grammatical coherence, the second question encouraged them to assess the preservation of meaning in the generated translation, a relatively relaxed assessment. {As shown in Table \ref{tab:human_eval_automated}, the annotators' responses to the first question indicate that the translation qualities were reliable. We observe high average scores on the Likert scale as well as strong inter-annotator agreements (computed using Cohen's $\kappa$) for all five languages. For the second question, the average scores on the Likert scale are consistently $\geq 4.10$ for all the five languages, indicating almost perfect preservation of meaning after translation from the English text to the target language.}

{Finally, we conducted a similar assessment of the quality of the human-translated subset of the Crisis Humanitarianism dataset. Each of 1000 (i.e., 200 $\times$ 5) translation pairs were similarity annotated by $3$ annotators. As expected, Table \ref{tab:human_eval_manual} shows that the fluency and meaning preservation in the human-translated text is better than the machine-translated text with strong inter-annotator agreement scores.}

In the upcoming sections, we describe the training and evaluation of the classification models, and the results for RQ1 and RQ2. Figure \ref{fig:method_overview} provides an overview of our method. 

\section{Language Disparity with Language Models}
\label{sec:language-specific}

In this section, we focus on RQ1: whether using large language models for classification tasks results in a systematic disparity between the performance on English and non-English text. We use pre-trained language models and fine-tune them on the specific classification task using language-specific labeled datasets.

\begin{table*}[!h]
    \centering
    \resizebox{\linewidth}{!}{
    \begin{tabular}{l | c  c  c  c | c  c  c  c |  c  c  c  c  }
        \textbf{Language - Model} & \multicolumn{4}{c|}{\textbf{Crisis Humanitarianism }} & \multicolumn{4}{c|}{\textbf{Fake News Detection} } & \multicolumn{4}{c}{\textbf{Emotion Detection}}\\
         &  $F_1$ & Precision & Recall & Accuracy  &  $F_1$ & Precision & Recall & Accuracy &  $F_1$ & Precision & Recall & Accuracy \\\midrule
         \textbf{Monolingual BERTs} & & & & & & & & & & & & \\
         \hspace{1mm} English - DisilBERT & 0.71 & 0.72 & 0.70 & 0.80 & 0.59 & 0.64 & 0.56 & 0.85 & 0.79 & 0.79 & 0.79 & 0.80\\
         \hspace{1mm} {Spanish - BETO} & {0.64} & {0.67} & {0.63} & {0.78} & {0.54} & {0.63} & {0.47} & {0.84}  & {0.75} & {0.76} & {0.75} & {0.77} \\
         \hspace{1mm} {French - CamemBERT} & {0.69} & {0.69} & {0.69} & {0.77} & {0.56} & {0.60} & {0.53} & {0.84} & {0.76} & {0.76} & {0.76} & {0.78} \\
         \hspace{1mm} {Portuguese - BERTimbau} &{0.67} & {0.67}  & {0.68} & {0.77} &{0.57} & {0.58} & {0.55} & {0.84} & {0.71} & {0.72} & {0.70} & {0.73} \\
         \hspace{1mm} {Chinese - ChineseBERT} & {0.65} & {0.64} & {0.66} & {0.75} & {0.56} & {0.61} & {0.51} & {0.84} & {0.72} & {0.72} & {0.72} & {0.74} \\
         \hspace{1mm} {Hindi - HindiBERT} & {0.63} & {0.62} & {0.64} & {0.74} & {0.54} & {0.59} & {0.51} & {0.83} &{0.70} & {0.71} & {0.69} & {0.71} \\\midrule
         \textbf{Multilingual BERT} & & & & & & & & & & & & \\
         \hspace{1mm} {English - mBERT} & {0.70} & {0.71} & {0.70} & {0.79} & {0.61} & {0.63} & {0.59} & {0.85} & {0.77} & {0.78} & {0.77} & {0.79} \\
         \hspace{1mm} Spanish  - mBERT & 0.62 & 0.65 & 0.61 & 0.77 & 0.57 & 0.59 & 0.55 & 0.84 &
         0.74 & 0.74 & 0.74 & 0.75 \\
         \hspace{1mm} French - mBERT & 0.68 & 0.68 & 0.69 & 0.77 & 0.58 & 0.59 & 0.56 & 0.84 &
         0.72 & 0.72 & 0.72 & 0.73\\
         \hspace{1mm} Portuguese - mBERT & 0.66 & 0.67 & 0.67 & 0.77 & 0.54 & 0.55 & 0.53 & 0.83 &
         0.71 & 0.71 & 0.71 & 0.72 \\
         \hspace{1mm} Chinese - mBERT & 0.62 & 0.61 & 0.64 & 0.74 & 0.54 & 0.60 & 0.49 & 0.84 &
         0.69 & 0.70 & 0.69 & 0.71 \\
         \hspace{1mm} Hindi - mBERT & 0.47 & 0.48 & 0.47 & 0.66 & 0.43 & 0.54 & 0.35 & 0.82 &
         0.64 & 0.65 & 0.64 & 0.67 \\
    \end{tabular}
    }
    \caption{{{Disparity between English and non-English languages using monolingual and multilingual models.}} Performance of the task and language-specific text-only classification models on $3$ datasets and $6$ languages.}
    \label{tab:modality_specific_results}
\end{table*}

\vspace{0.05in} \noindent\textbf{Classification models for English:} {We use two pre-trained language models:  DistilBERT~\cite{sanh2019distilbert} (\texttt{distilbert-base-cased} on HuggingFace) and DistilmBERT (\texttt{distilbert-base-multilingual-cased} on HuggingFace) to classify the English text.} We fine-tune the pre-trained language models on the 3 datasets discussed above by using the respective training sets. The process of fine-tuning a language model involves taking a pre-trained language model\footnote{Large language models are typically pre-trained in a self-supervised manner (e.g., predicting a masked word, given other surrounding words that contextualize the masked word~\cite{DBLP:journals/corr/abs-1810-04805}) using corpora that comprises billions of words.} and replacing the ``pre-training head'' of the model with a randomly initialized ``classification head''. The randomly initialized parameters in the classification head are learned by \textit{fine-tuning} the model on classification examples while minimizing the cross-entropy loss. To train the English language classification models for each dataset, we use Adam optimizer~\cite{kingma2014adam} with a learning rate initialized at $10^{-4}$; hyper-parameters are set by observing the classification performance achieved on the respective validation set. We use early stopping~\cite{caruana2000overfitting} to stop training when the loss value on the validation set stops to improve for 5 consecutive epochs.

\vspace{0.05in} \noindent\textbf{Classification models for non-English languages:} {To classify the non-English languages into task-specific categories, we use two set of pre-trained language models: \textit{(a)} monolingual models and \textit{(b)} multilingual model called DistilmBERT (\texttt{distilbert-base-multilingual-cased} on HuggingFace). For monolingual models, we refer to the leaderboard maintained by \citeauthor{nozza2020mask} (\citeyear{nozza2020mask}) and select the best performing models for a specific language. Namely, we select BETO for modeling Spanish text~\cite{CaneteCFP2020}, CamemBERT for French~\cite{martin-etal-2020-camembert}, BERTimbau for Portuguese~\cite{souza2020bertimbau}, ChineseBERT for Chinese~\cite{cui-etal-2020-revisiting}, and HindiBERT for Hindi~\cite{HindiBERT}. We adopt the same model training and hyper-parameter selection strategies as for the English language models discussed above. Training a classification model for each of the five non-English languages across the three tasks gives us a total of $30$ non-English text classification models.} Our training strategies allow us to compare the \textit{best} text classification models for all the languages for each of the three tasks individually.

\vspace{0.05in} \noindent\textbf{Fine-tuned text representations}: Once fine-tuned, the text classifiers can be used to extract representations for any input text by taking the output of the penultimate layers. These representations, also called \textit{embeddings}, capture attributes of the text that the model has learned to use for categorizing the input into the target classes, and therefore can be fed to the multimodal classifier as a representation of the text part of the multimodal input. We obtain this latent representation of input text, denoted by vector $\mathbf{T}$ (with dimension $768$), by averaging the token-level outputs from the penultimate layer of the fine-tuned classification models.

\vspace{0.05in} \noindent\textbf{Evaluation metrics}: We compute standard classification metrics to evaluate the performance these text-only classifiers on the test sets of respective datasets. Since crisis humanitarian post detection and emotion detection are multi-class classification tasks, we compute macro averages of class-wise $F_1$, precision, and recall scores along with the overall classification accuracy. However, since fake news detection is a binary classification task, we compute the $F_1$, precision, and recall scores for the positive class (i.e., \texttt{label:1} $=$ \texttt{fake}). Table \ref{tab:modality_specific_results} summarizes the performance of the text-only classifiers discussed above.
Since the performance of deep learning models, especially BERT-based large language models, can possibly change with initialization schemes~\cite{sellam2021multiberts}, we vary the random initialization across different runs of the models and report the averages from $10$ different runs. 

\vspace{0.05in} \noindent\textbf{Performance on English vs. non-English languages:} In Table \ref{tab:modality_specific_results}, we observe that the performance of text-only classification models is higher when the input is in the English language when compared against the performance of models that take other high-resource non-English languages as input. {This trend is consistent across \textit{(i)} both monolingual and multilingual models, \textit{(ii)} the three tasks considered in this work as well as \textit{(iii)} across all the classification metrics. For monolingual and multilingual models, the gap in performance on English and non-English languages varies with the task at hand as well as the non-English language being considered. For instance, for the crisis humanitarianism task with monolingual models, the drop in $F_1$ score of Spanish with respect to that of English is $9.5\%$, while it is $5.1\%$ for the emotion detection task. For the same task, e.g., emotion detection, using monolingual models leads to performance drops that vary from $5.1\%$ for Spanish to $11.4\%$ for Hindi. It is noteworthy that the performance on non-English languages relative to each other maintains a near-uniform pattern across the three tasks for both monolingual and multilingual models -- the performance is consistently the worst for Hindi; the performance on Chinese and Portuguese is relatively better, and the performance on Spanish and French is best when compared against other non-English languages. We revisit this observation and its potential causes in the Discussion section. In sum, our results indicate a language disparity exists due to the use of large language models in varied classification tasks --- whether monolingual or multilingual.} We recall that the adopted methodology -- fine-tuning of pre-trained language models -- is representative of the state-of-the-art NLP techniques that are frequently adopted for solving classification tasks~\cite{li2020survey}.

\begin{table*}[!h]
    \centering
    \resizebox{\linewidth}{!}{
    \begin{tabular}{l | c  c  c  c | c  c  c  c | c  c  c  c  }
        \textbf{Input} & \multicolumn{4}{c|}{\textbf{Crisis Humanitarianism }} & \multicolumn{4}{c|}{\textbf{Fake News Detection} } & \multicolumn{4}{c}{\textbf{Emotion Detection}}\\
         &  $F_1$ & Precision & Recall & Accuracy  &  $F_1$ & Precision & Recall & Accuracy &  $F_1$ & Precision & Recall & Accuracy \\\midrule
         Image-only & 0.42 & 0.45 & 0.42 & 0.52 & 0.15 & 0.54 & 0.09 & 0.81 & 0.94 & 0.94 & 0.94 & 0.95\\\midrule
         \textbf{Monolingual BERTs} & & & & & & & & & & & & \\
         \hspace{1mm} English + Image & 0.73 & 0.74 & 0.72 & 0.82 & 0.60 & 0.63 & 0.58 & 0.85 &
         0.85 & 0.87 & 0.84 & 0.86\\
         \hspace{1mm} {Spanish + Image} & {0.72} & {0.73} & {0.71} & {0.82} & {0.59} & {0.63} & {0.57} & {0.85} & {0.82} & {0.83} & {0.81} & {0.82} \\
         \hspace{1mm} {French + Image} & {0.71} & {0.72} & {0.69} & {0.81} & {0.58} & {0.61} & {0.55} & { 0.84} & {0.81} & {0.82} & {0.81} & {0.82} \\
         \hspace{1mm} {Portuguese + Image} & {0.71} & {0.71} & {0.70} & {0.80} & {0.59} & {0.60} & {0.58} & {0.84} & {0.81} & {0.82} & {0.81} & {0.82} \\
         \hspace{1mm} {Chinese + Image} & {0.70} & {0.69} & {0.70} & {0.80} & {0.58} & {0.62} & {0.54} & {0.84} & {0.80} & {0.80} & {0.79} & {0.81} \\
         \hspace{1mm} {Hindi + Image} & {0.68} & {0.69} & {0.67} & {0.80} & {0.56} & {0.61} & {0.51} & {0.84} & {0.78} & {0.79} & {0.77} & {0.80} \\\midrule
         \textbf{Multilingual BERT} & & & & & & & & & & & & \\
         \hspace{1mm} {English + Image} & {0.75} & {0.78} & {0.73} & {0.82} & {0.61} & {0.63} & {0.60} & {0.85} & {0.80} & {0.82} & {0.79} & {0.82} \\
         \hspace{1mm} Spanish + Image & 0.75 & 0.77 & 0.74 & 0.81 & 0.60 & 0.64 & 0.56 & 0.85 &
         0.76 & 0.80 & 0.75 & 0.76 \\
         \hspace{1mm} French + Image & 0.74 & 0.84 & 0.71 & 0.83 & 0.58 & 0.60 & 0.57 & 0.84 &
         0.76 & 0.76 & 0.76 & 0.77 \\
         \hspace{1mm} Portuguese + Image & 0.76 & 0.76 & 0.76 & 0.82 & 0.56 & 0.55 & 0.57 & 0.83 &
         0.77 & 0.77 & 0.78 & 0.79 \\
         \hspace{1mm} Chinese + Image & 0.73 & 0.75 & 0.71 & 0.80 & 0.55 & 0.52 & 0.57 & 0.83 &
         0.77 & 0.79 & 0.76 & 0.79 \\
         \hspace{1mm} Hindi + Image & 0.64 & 0.68& 0.61 & 0.78 & 0.46 & 0.57 & 0.38 & 0.83 &
         0.75 & 0.76 & 0.74 & 0.76 \\
    \end{tabular}
    }
    \caption{{Image-only and multimodal classfication performance.} Performance of task-specific image-only classifiers (Row 1) and task- and language-specific multimodal classifiers (both monolingual and multilingual). }
    \label{tab:multimodal_results}
\end{table*}

\section{Benefits of Multimodal Learning}
In this section, we focus on RQ2: can we leverage images with the help of multimodal learning to overcome the disparity between English and non-English languages. 

\vspace{0.05in} \noindent\textbf{Image-Only classification model:} To investigate the predictive power of images without textual information, we develop and evaluate image-only classifiers for each dataset. Similar to text classifiers, we apply a fine-tuning approach to train the task-specific image classifiers. We first freeze the weights of VGG-16~\cite{simonyanVeryDeepConvolutional2015a}, a popular deep Convolutional Neural Network, pre-trained on ImageNet~\cite{imagenet2009Deng}, a large-scale generic image classification dataset. Then, we swap the last layer from the original model to three fully connected hidden layers with dimensions \texttt{4096}, \texttt{256}, and \texttt{num-of-classes}, respectively.
Finally, retrain these three layers to adapt the image distribution in each dataset.

\begin{figure*}[!ht]
   \subfloat[\label{crisis_plot}][Crisis Humanitarianism]{%
      \includegraphics[ width=0.33\textwidth]{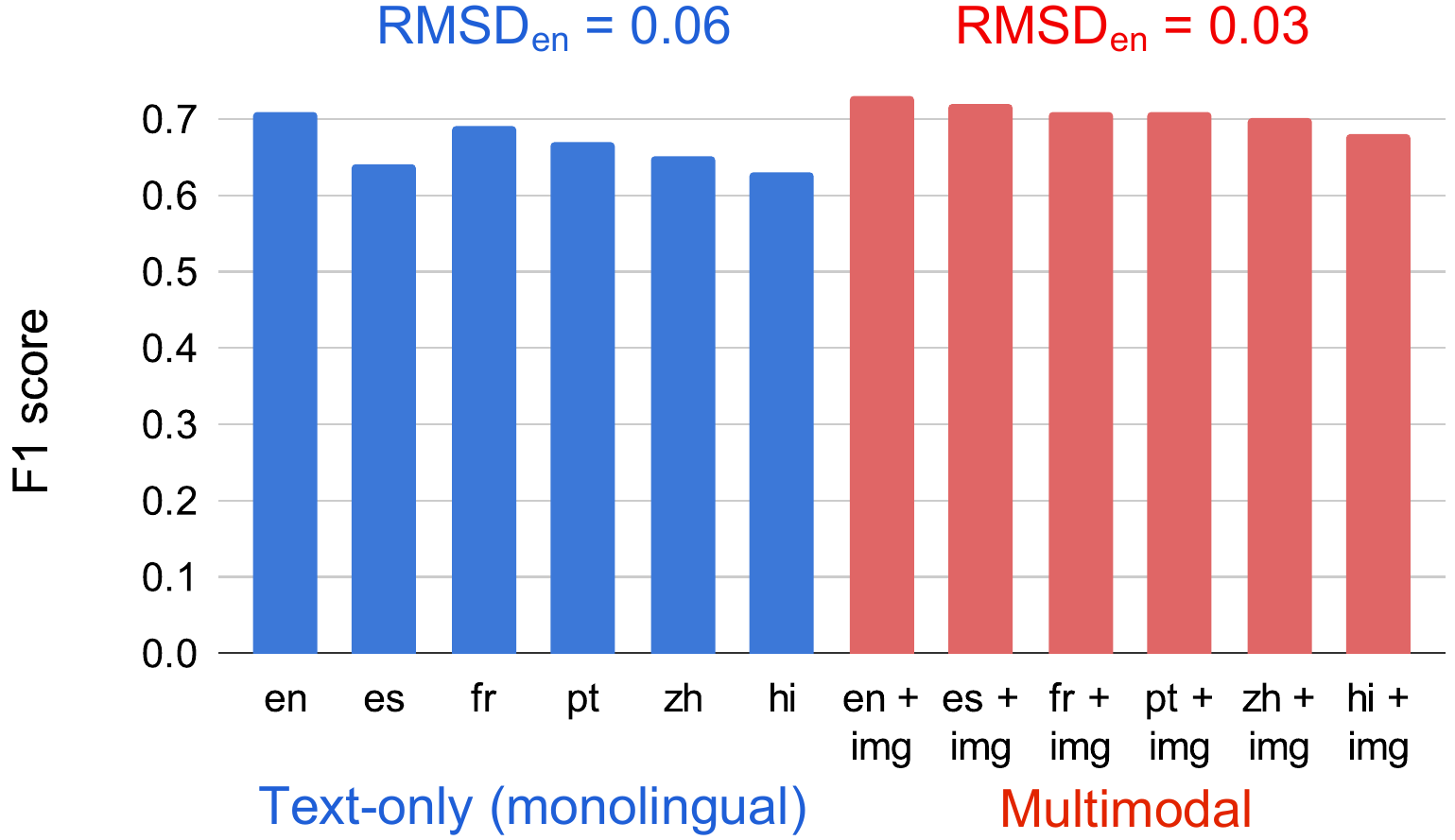}}
\hspace{\fill}
   \subfloat[\label{fakenews_plot}][Fake News Detection]{%
      \includegraphics[ width=0.33\textwidth]{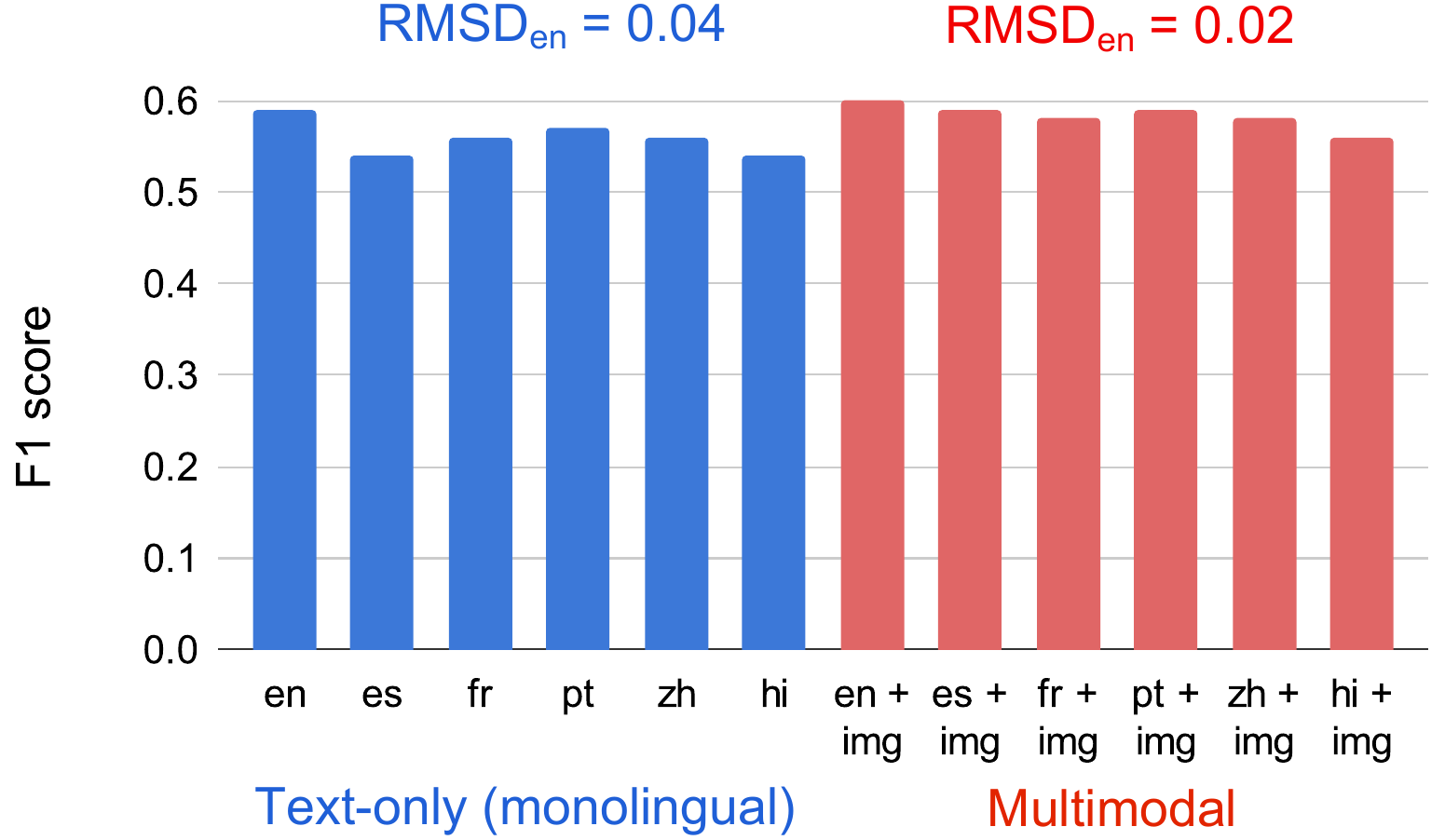}}
\hspace{\fill}
   \subfloat[\label{emotion_plot}][Emotion Detection]{%
      \includegraphics[ width=0.33\textwidth]{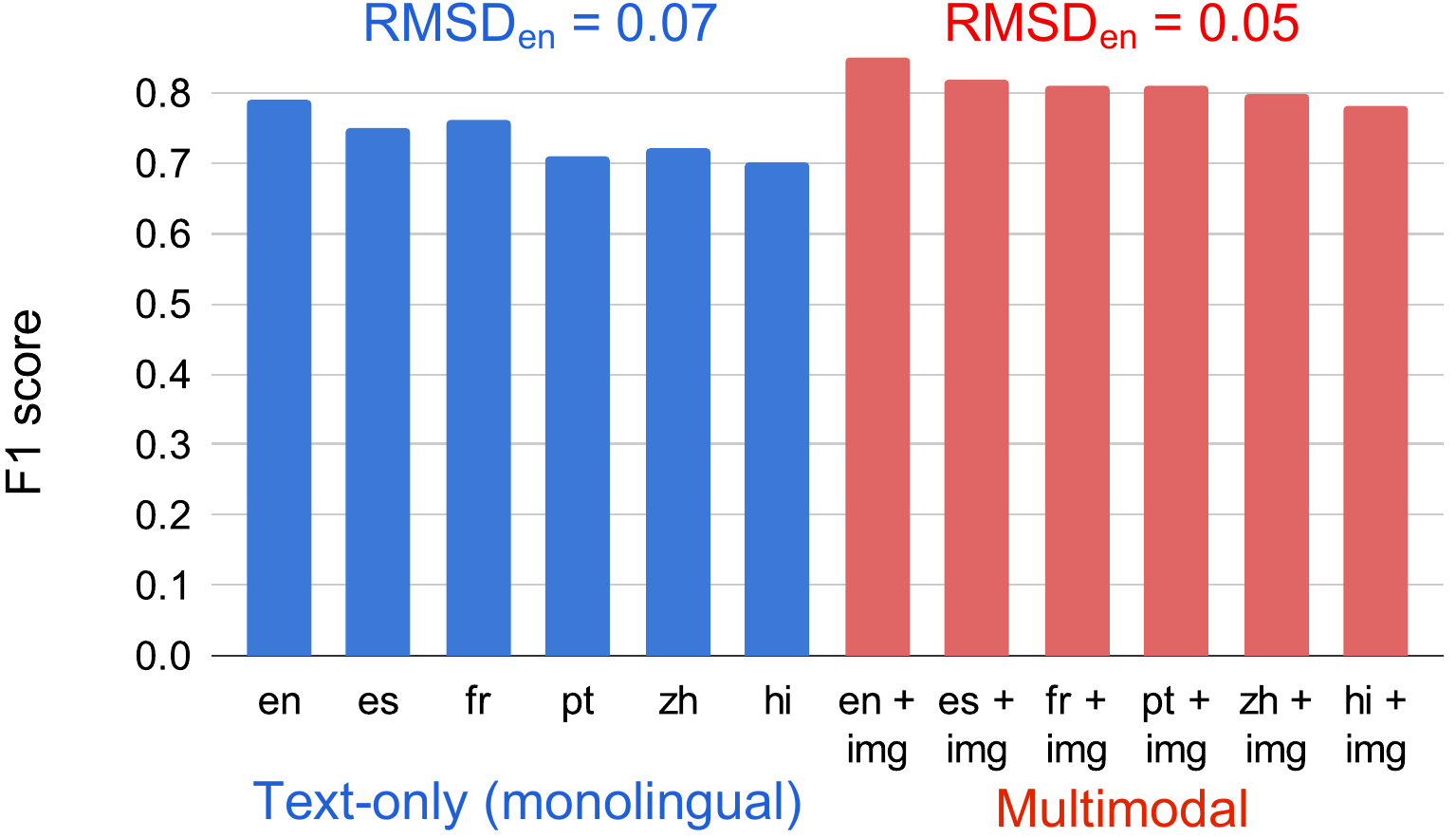}}\\
\caption{{{Comparing $F_1$ scores on non-English and English text for both text-only and multimodal classifiers using \textit{monolingual} language models.} RMSD$_{en}$ denotes the root-mean-square deviation of the $F_1$ scores achieved by non-English classifiers  with respect to the that of the corresponding English classifier. The RMSD$_{en}$ values for multimodal models are lower than those for monolingual text-only models.}}
\label{fig:rmsd_plots_monolingual}
\end{figure*}

\begin{figure*}[!ht]
   \subfloat[\label{crisis_plot}][Crisis Humanitarianism]{%
      \includegraphics[ width=0.33\textwidth]{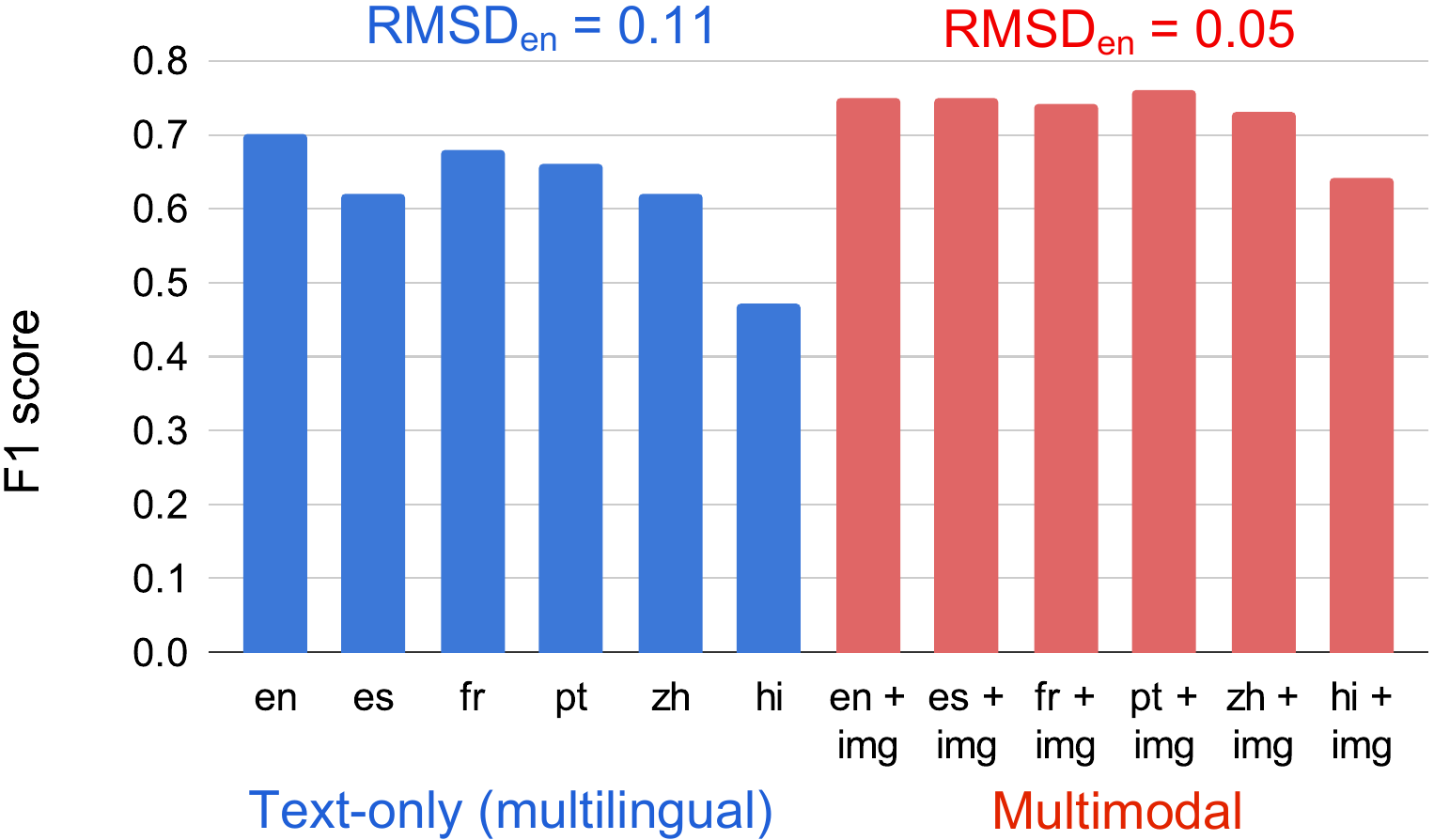}}
\hspace{\fill}
   \subfloat[\label{fakenews_plot}][Fake News Detection]{%
      \includegraphics[ width=0.33\textwidth]{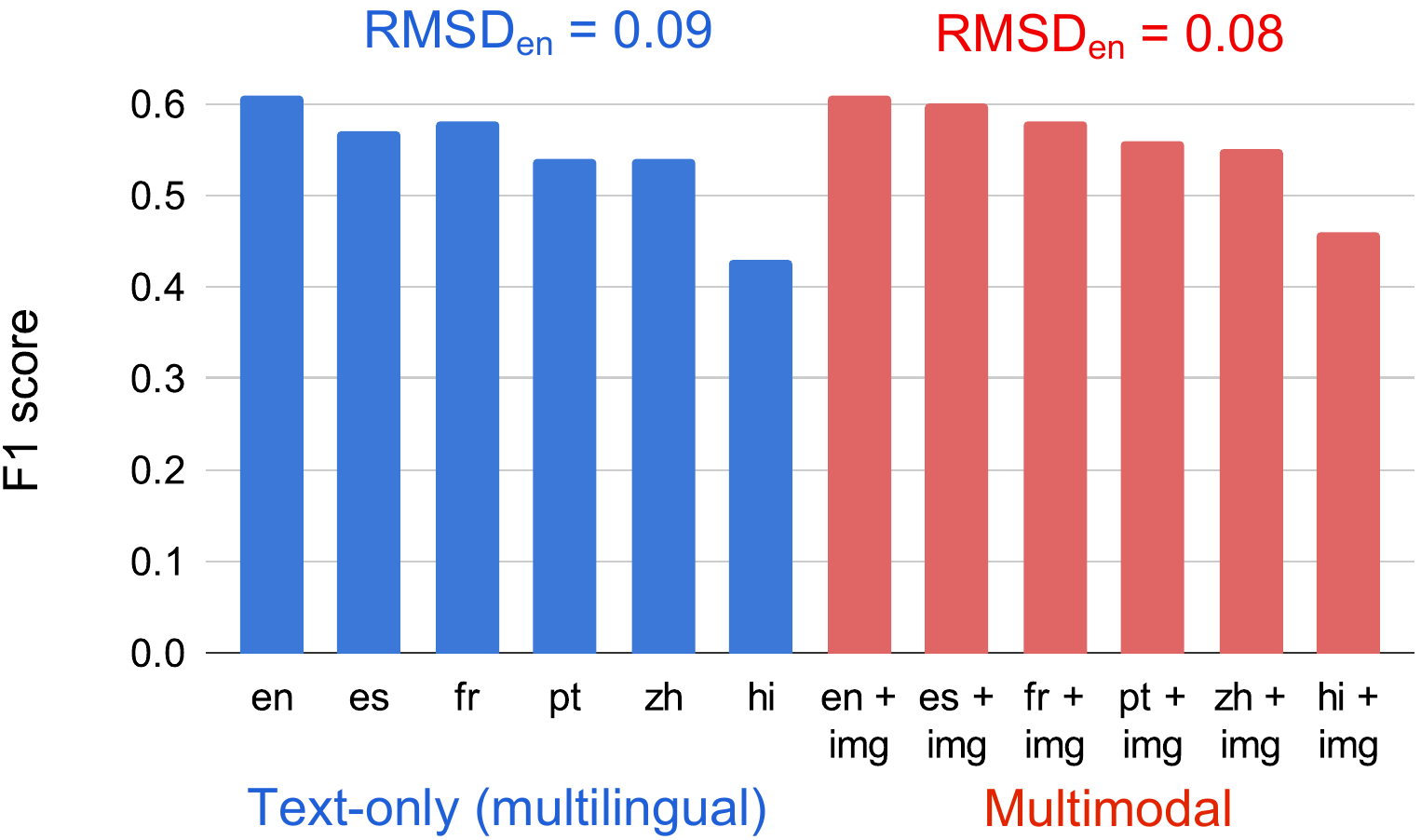}}
\hspace{\fill}
   \subfloat[\label{emotion_plot}][Emotion Detection]{%
      \includegraphics[ width=0.33\textwidth]{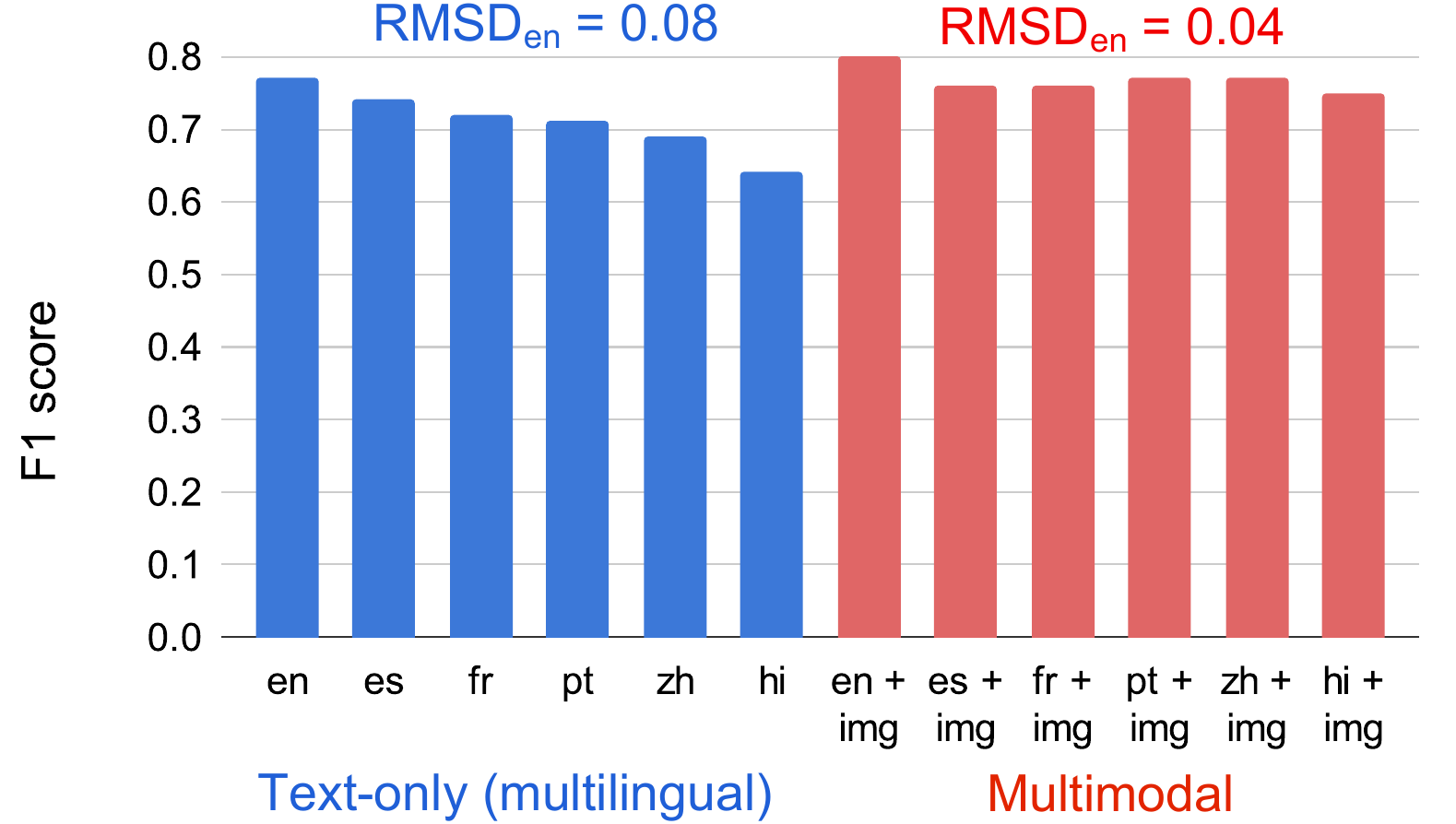}}\\
\caption{{{Comparing $F_1$ scores on non-English and English text for both text-only and multimodal classifiers using \textit{multilingual} models.} RMSD$_{en}$ denotes the root-mean-square deviation of the $F_1$ scores achieved by non-English classifiers  with respect to the that of the corresponding English classifier. The RMSD$_{en}$ values for multimodal models are lower than those for multilingual text-only models.}}
\label{fig:rmsd_plots_multilingual}
\end{figure*}

As images in our datasets have various dimensions, we apply a standard image pre-processing pipeline so that they can fit the pre-trained VGG-16 model's input requirement.
We first resize the image so that its shorter dimension is $224$, then we crop the square region in the center and normalize the square image with the mean and standard deviation of the ImageNet images~\cite{imagenet2009Deng}.

To train and evaluate image-specific classifiers, we use the same splits in text-only models to divide images into the train, validation, and test sets.
We use Adam optimizer~\cite{kingma2014adam} with a learning rate of $10^{-4}$ for each dataset.
To avoid overfitting, we use early stopping to stop training when the loss value on the validation set stops to improve for 10 consecutive epochs. 
Finally, we extract the image embeddings, denoted by $\mathbf{I}$, from image-specific classifiers by computing the neuron activations from the penultimate layer (with dimension $256$) as a latent representation of the image information for our multimodal models.

\vspace{0.05in} \noindent\textbf{Multimodal classification model:}
We implement a multimodal classifier~\cite{ngiam2011multimodal} that fuses the latent representations of individual modalities (text and image) to perform classification based on the joint modeling of both input modalities. We feed the concatenation of fine-tuned text and image representations (i.e., $\mathbf{T} \oplus \mathbf{I}$) to the multimodal classifier, which is essentially a series of fully connected layers with ReLU activation~\cite{agarap2018deep}. The architecture of the multimodal classifier comprises an input layer ($1024$ neurons), $3$ hidden layers ($512$, $128$, $32$ neurons), and an output layer (neurons $=$ number of classes in the dataset). We train a multimodal classifier for each language in each task. Similar to image-only and text-only classification models discussed above, for each training instance, we use Adam optimizer~\cite{kingma2014adam} with a learning rate initialized at $10^{-4}$. We use early stopping based on the validation set loss to stop the training and avoid overfitting on the train set. 

We use the same evaluation metrics to evaluate the image-only and multimodal classifiers as we did for the text-only ones, and report the average of $10$ different runs in Table \ref{tab:multimodal_results}. {Additionally, in Figures \ref{fig:rmsd_plots_monolingual} and \ref{fig:rmsd_plots_multilingual} we present the root-mean-squared deviation (RMSD$_{en}$) values of $F_1$ scores of non-English languages with respect to that of the English language for text-only and multimodal classifiers.}

\vspace{0.05in} \noindent\textbf{Multimodal learning boosts classification performance:} As Table \ref{tab:multimodal_results} shows, the classification performance for all the languages (English as well as non-English) improves considerably with the inclusion of images as an additional modality when compared against the performance of corresponding text-only classification models. {This trend is consistent across all three datasets and both the set of models considered in our study --- monolingual as well as multilingual.}
It is interesting to note that the benefit of including images, as indicated by the increase in performance metrics, is largely dependent on the \textit{informativeness} of the images towards the classification task. For instance, for fake news detection, the image-only classifier achieves an $F_1$ score of $0.15$, indicating poor distinguishability between real and fake news solely based on images in a news article. {Consequently, the increase in the performance of the multimodal classifier over that of the monolingual text-only classifier is relatively marginal, ranging from $1.5\%$ ($F_1$ increases from $0.59$ to $0.60$ for English) to $3.7\%$ ($F_1$ increases from $0.54$ to $0.56$ for Hindi).} In contrast, for the emotion detection task, the image-only classifier achieves an $F_1$ score of $0.94$, indicating extremely good distinguishability between emotion categories solely based on images. {As a consequence, the increase in the performance of the multimodal classifier over that of the monolingual text-only classifier ranges from $7.6\%$ ($F_1$ increases from $0.79$ to $0.85$ for English) to $11.4\%$ ($F_1$ increases from $0.70$ to $0.78$ for Hindi). We observe the same trends for multilingual models as well.} 

\vspace{0.05in} \noindent\textbf{Multimodal learning helps in bridging the gap between English and non-English languages:} The results discussed so far indicate: \textit{(i)} the performance of the state-of-the-art techniques for non-English languages is worse than the performance of the state-of-the-art techniques for the English language, and \textit{(ii)} incorporating images as an additional modality using multimodal learning leads to better classification performance when compared against the performance of  text-only counterparts. However, a crucial question remains to be answered: can multimodal learning help in overcoming the language disparity between English and non-English languages? To answer this, we focus on the root-mean-square deviation (RMSD$_{en}$) scores presented in Figures \ref{fig:rmsd_plots_monolingual} and \ref{fig:rmsd_plots_multilingual}. RMSD$_{en}$ is calculated by taking the root of the average of the squared pairwise differences between $F_1$ scores for English and other non-English languages. {We compute the RMSD$_{en}$ scores for both monolingual and multilingual models.}  It is clear that the RMSD$_{en}$ of $F_1$ scores achieved by non-English classifiers with respect to the $F_1$ score achieved by the English classifier are lesser with multimodal input when compared against text-only input. {For monolingual models, the drops in RMSD$_{en}$ values are $50.0\%$ ($0.06 \rightarrow 0.03$; Figure \ref{fig:rmsd_plots_monolingual}(a)), $50.0\%$ ($0.04 \rightarrow 0.02$; Figure \ref{fig:rmsd_plots_monolingual}(b)), and $28.6\%$ ($0.07 \rightarrow 0.05$; Figure \ref{fig:rmsd_plots_monolingual}(c)) for crisis humanitarianism, fake news detection, and emotion detection, respectively. Similarly, for the multilingual models, the drops in RMSD$_{en}$ values are $54.5\%$ ($0.11 \rightarrow 0.05$; Figure \ref{fig:rmsd_plots_multilingual}(a)), $11.1\%$ ($0.09 \rightarrow 0.08$; Figure \ref{fig:rmsd_plots_multilingual}(b)), and $50.0\%$ ($0.08 \rightarrow 0.04$; Figure \ref{fig:rmsd_plots_multilingual}(c)) for crisis humanitarianism, fake news detection, and emotion detection, respectively.  The drop in deviation with respect to the scores for English demonstrates that images are effective in bridging the gap between English and non-English languages. This is also pictorially depicted in Figures \ref{fig:rmsd_plots_monolingual} and \ref{fig:rmsd_plots_multilingual}, as the red bars (with multimodal input) are more uniform in length than the blue bars (with text-only input).}

\begin{table}[!h]
    \centering
    \resizebox{6cm}{!}{
    \begin{tabular}{l | c  c }
        \textbf{{Language \& Model}} & \multicolumn{2}{c}{\textbf{{Crisis Humanitarianism}}}\\
        & \textbf{{Language-only}} & \textbf{{Multimodal}}\\
        & {$F_1$ score} & {$F_1$ score}\\\midrule
        \textbf{{Monolingual BERTs}} & & \\ 
        \hspace{1mm} {English} & {0.68} & {0.72}  \\
        \hspace{1mm} {Spanish} & {0.63} & {0.69} \\
        \hspace{1mm} {French} & {0.64} & {0.70} \\
        \hspace{1mm} {Portuguese} & {0.63} & {0.68} \\
        \hspace{1mm} {Chinese} & {0.64} & {0.67} \\
        \hspace{1mm} {Hindi} & {0.61} & {0.66} \\\midrule
        \textbf{{Multilingual BERT}} & & \\
        \hspace{1mm} {English} & {0.69} & {0.73} \\
        \hspace{1mm} {Spanish} & {0.62} & {0.72} \\
        \hspace{1mm} {French} & {0.63} & {0.72} \\
        \hspace{1mm} {Portuguese} & {0.61} & {0.69} \\
        \hspace{1mm} {Chinese} & {0.60} & {0.66} \\
        \hspace{1mm} {Hindi} & {0.44} & {0.61} \\\midrule
    \end{tabular}
    }
    \caption{{{Classification performance on \textit{human} translated crisis humanitarianism test set.} Performance of language-only and multimodal classifiers. The reported values are averages of 10 different runs.}}
    \label{tab:results_human_translated_data}
\end{table}

\vspace{0.05in} \noindent\textbf{{Results on human-translated test set:}} {To evaluate the performance of trained models on a sample that is free from the noise introduced by automated translators, we evaluate all the trained models for the crisis humanitarian task on the human-translated subset of the test set. Table \ref{tab:results_human_translated_data} reinforces our observations --- the disparity between English and non-English languages exists due to both monolingual and multilingual language models and multimodal learning helps in reducing this performance gap. For monolingual and multilingual models, the RMSD$_{en}$ values drop from $0.05$ to $0.04$ and from $0.15$ to $0.06$, respectively.}

\section{Discussion}

Our study demonstrates that in the context of societal tasks -- as demonstrated by our focus on three datasets -- the performance of large language models on non-English text is subpar when compared to the performance on English text. 
In the subsequent discussion, we highlight how this could have possibly threatening implications on the lives of many individuals who belong to underserved communities. 

Furthermore, we empirically demonstrate that using images as an additional modality leads to a lesser difference between the performance on English and non-English text, as indicated by decreased RMSD$_{en}$ values. While existing studies have focused on developing advanced monolingual language models that can boost the performance on specific non-English languages to bridge the performance gap, we demonstrate the benefits of including other complementary modalities, especially those that are language-agnostic. Decreased RMSD$_{en}$ values indicate that if images are considered along with the text, the performance on all languages is not only better than when \textit{only text} is considered, but it is also comparable across English and non-English languages.

\vspace{0.05in} \noindent\textbf{Implications of language disparity with text-only models:} 
In the context of social computing, disparities between English and non-English languages can lead to inequitable outcomes. 
For instance, as per our observations, if state-of-the-art NLP techniques that are centered around BERT-based language models are adopted to detect humanitarian information during crises, the detection abilities would be poorer for social media posts in non-English languages than those in English, causing delayed interventions.
In countries like the United States, where non-English languages like Spanish and Chinese are spoken by a considerable number of people~\cite{american2016state}, this disparity could exacerbate the effects of discrimination and prejudice that they already face~\cite{pew_latinos}. Similarly, poor emotion recognition in specific non-English languages can lead to unhelpful or even harmful outcomes in scenarios where the output of emotion recognition informs mental health interventions. Furthermore, poor fake news detection in specific non-English languages can lead to lacking correction and mitigation efforts, leading to relatively worse outcomes for non-English speaking populations.

\vspace{0.05in} \noindent\textbf{Implications of reduced language disparity with multimodal models:} 
People use multiple content modalities -- images, text, videos, and audio clips, to share updates on social platforms. Visual modalities (like images and videos) transcend languages and are extremely informative in scenarios like crisis information detection and emotion detection. Combining our multimodal approach with existing text-only approaches for better modeling of non-English text can present \textit{complementary gains}, leading to a reduced gap between English and non-English languages. In other words, an approach that complements existing approaches that focus on only text can be expected to provide gains even as the language-specific text-only approaches improve and evolve. 

\begin{figure}[!t]
   \subfloat[\label{text_only_pretraining_size}][Text-only models]{%
      \includegraphics[ width=0.495\linewidth]{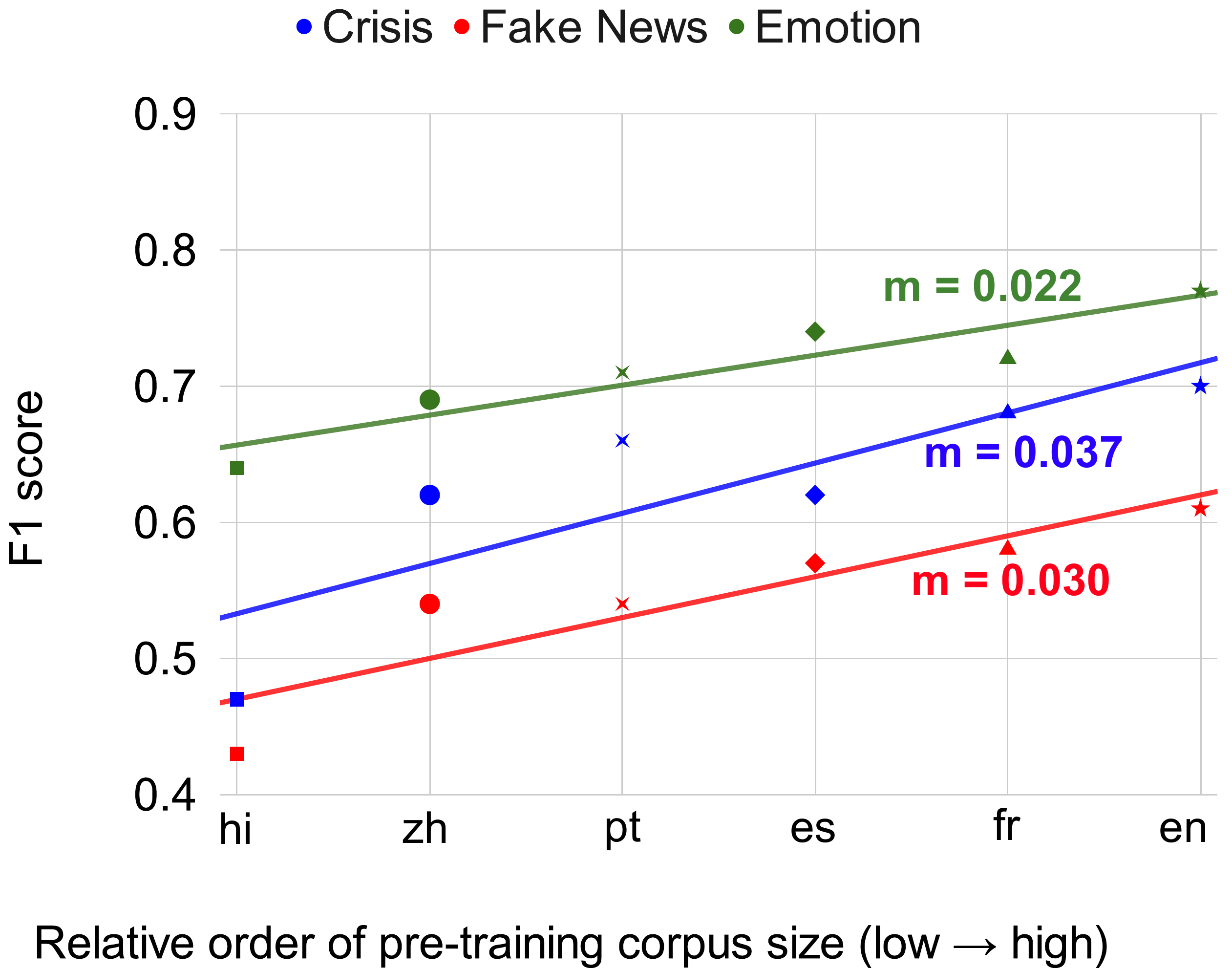}}
\hspace{\fill}
   \subfloat[\label{multimodal_pretraining_size}][Multimodal models]{%
      \includegraphics[ width=0.495\linewidth]{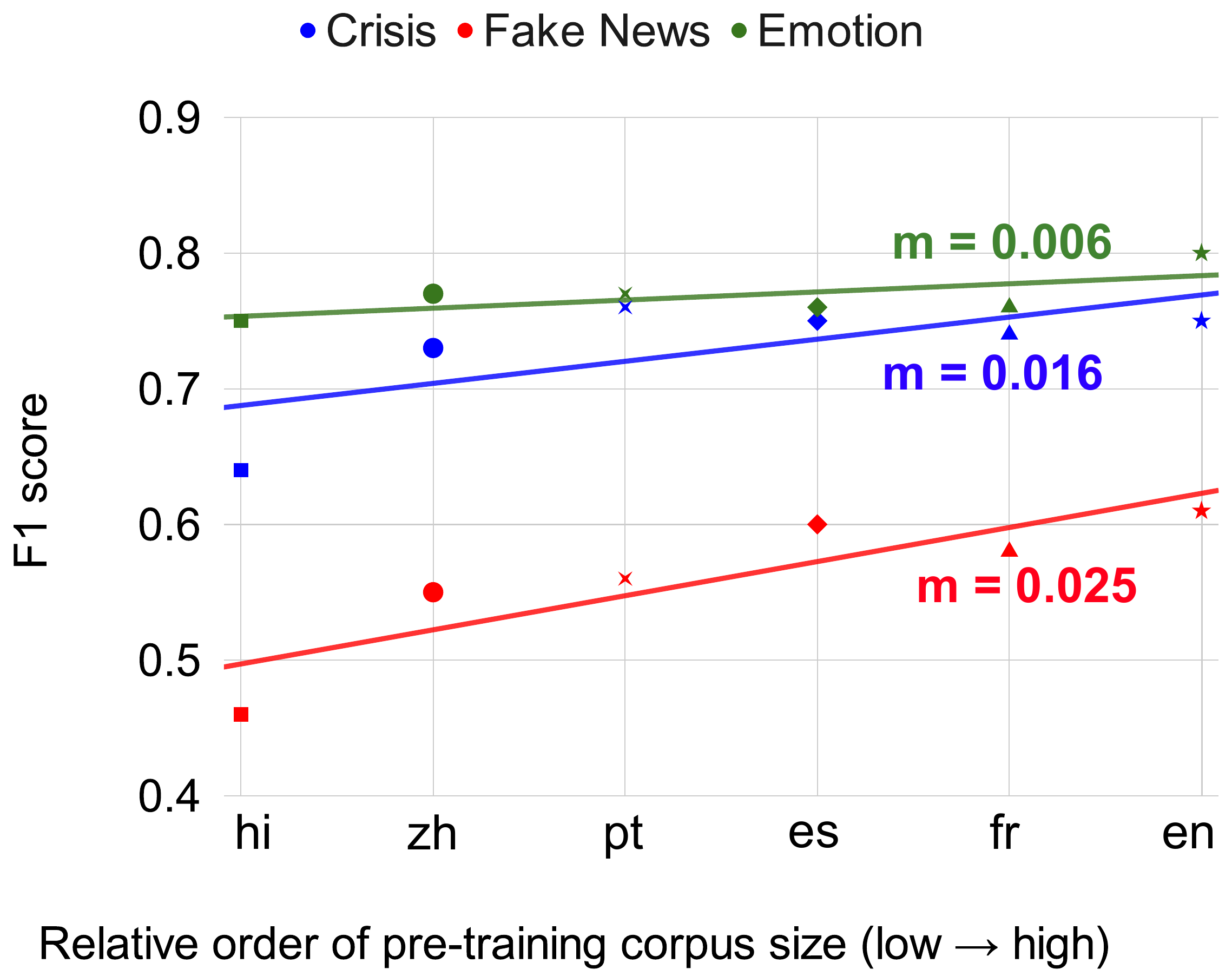}}\\
\caption{{{Relation between pre-training corpus size and classification performance.} The languages on the x-axes are ordered as per their representation in the pre-training corpora of mBERT; y-axes report the $F_1$ scores achieved on all the considered languages and task after task-specific fine-tuning. $m$ denotes the slope of the task-wise trend lines.} }
\label{fig:pre_training}
\end{figure}

\vspace{0.05in}\noindent\textbf{Dependence of performance on pre-training corpus size:} {The multilingual language model used in this study --- mBERT, was pre-trained on huge corpora using self-supervised objectives~\cite{DBLP:journals/corr/abs-1810-04805}.} 
The data sizes (in GiB) in mBERT's pre-training corpus have the relative order \texttt{en} $>$ \texttt{fr} $>$ \texttt{es} $>$ \texttt{pt} $>$ \texttt{zh} $>$ \texttt{hi}~\cite{conneau2020unsupervised}. As shown in Figure \ref{fig:pre_training}(a), the relationship between the language-specific corpus size that mBERT is trained on and the classification performance obtained after task-specific fine-tuning, is clear: larger representation in the pre-training corpus is related to better performance on downstream tasks. This trend reinforces the findings of ~\citeauthor{wu2020all} (\citeyear{wu2020all}) in our context --- the performance of large language models drops significantly as the considered languages have less pre-training data. This is concerning because, as \citeauthor{bender2021dangers} (\citeyear{bender2021dangers}) argue, ``the large, uncurated, and Internet-based datasets'' that these language models are trained on ``encode the dominant/hegemonic view, which further harms people at the margins.''
However, as shown in Figure \ref{fig:pre_training}(b), incorporating images using multimodal learning leads to a weakened dependence on the pre-training corpus size. This is indicated by the reduced slopes ($m$) of the trend lines across all three tasks. In effect, we demonstrate that multimodal learning, if adopted in the fine-tuning stages of approaches that employ large language models, could help in overcoming the well-recognized dependence of downstream performance on language-specific pre-training corpus size.

Beyond the theoretical implications discussed above, we believe our methods demonstrate the crucial role that multimodal learning can play in the equitable dissemination of NLP-based services to a broader range of the global population. The systems that make inferences based on text data alone can be adapted to include the information contained in images, wherever possible, leading to better detection abilities on the non-English text and thereby bridging the gap between English and non-English languages. {As our evaluation on human-translated and machine-translated text demonstrates, our proposed approach is compatible with setups that infer information directly from non-English text and with the approaches that first translate non-English text to English and then infer information from the translations.}

\vspace{0.05in}\noindent\textbf{Limitations and future work:} 
Large language models such as T5 and their corresponding multilingual variants mT5 overcome several limitations of BERT and mBERT by adopting different pre-training strategies. We specifically focused on BERT-based language models as representatives of large language models -- note that our study aimed to understand the effectiveness of multimodal learning in overcoming the language disparity and not the relative performance of different language models. 
Since the underlying idea of fusing image and text representations can be applied to other language models as well, we believe that our insights and takeaways will also generalize to them.

In the future, we intend to experiment with low-resource languages to expand our claims to a wider set of languages. There are two major challenges on those fronts: \textit{(i)} availability of parallel data, and \textit{(ii)} identifying and developing the state-of-the-art text-only classification approaches for low-resource languages. A translation-based data creation pipeline will not work for low-resource languages and hence we may either curate the data by recruiting native speakers to translate the original examples from English or by collecting real data from social media for different languages. Furthermore, since the state-of-the-art classification approach for low resource languages may not be based on large language models~\cite{wu2020all, nozza2020mask}, we intend to identify and develop those language-specific approaches. 

Lastly, the current study focuses on bridging the gap that exists in classification tasks. As part of future work, we intend to explore other types of tasks that are relevant to the social computing theme. Such tasks include, analyzing the lifestyle choices of social media users~\cite{islam2021analysis} and context-based quotation recommendation~\cite{maclaughlin2021context}. By including other modalities like images, these approaches may be extended to non-English speaking populations. However, while images are not bound by languages, their production and perception are culturally influenced~\cite{hong2003boundaries}. This cultural influence is more prominent in user-generated content that is abundant on social platforms~\cite{shen2019measuring}. Therefore, it is important to consider the cultural confounds in the production and consumption of images while using them to train and infer from machine learning models.

\vspace{0.05in}\noindent\textbf{{Broader perspective, ethics, and competing interests:}} {Developing powerful, accessible, and equitable resources for modeling non-English languages remains an open challenge. Our work argues that including information from other modalities, specifically images, can present new avenues to progress research in this direction. We believe this work will positively impact society by motivating researchers and practitioners to develop more reliable classifiers for non-English languages with applications to societal tasks. That said, it is worth noting that since images alone do not represent the entire cultural context, modeling techniques for non-English languages should continue to develop. Incorporation of new modalities alongside text also comes with additional challenges --- for instance, the biases that computer vision models encode~\cite{hendricks2018women} need to be taken into consideration, and methods need to be developed to model cultural shifts in meaning for similar images~\cite{liu2021visually}. The authors involved in this study do not have any competing interests that could have influenced any part of the conduct of this research.}

\section{Conclusion}
In sum, we have demonstrated that the adoption of large language models for building approaches for tasks aimed at detecting humanitarian information, fake news, and emotion leads to systematically lower performance on non-English languages when compared to the performance on English. We discussed how such a disparity could lead to inequitable outcomes. Furthermore, we empirically show that including images via multimodal learning bridges this performance gap. Our experiments yield consistent insights on $3$ different datasets and $5$ non-English languages, indicating their generalizability. We also discussed the reliance of large language models on pre-training corpus size and how adopting multimodal learning during fine-tuning stages can weaken this dependence, leading to a more consistent performance across all languages under consideration.

\section{Acknowledgements}
This research has been supported in part by NSF IIS-2027689, NSF ITE-2137724, Microsoft AI for Health, and IDEaS at Georgia Tech. We thank Sindhu Kiranmai Ernala, Sandeep Soni, and Talayeh Aledavood for helpful discussions in the early stages of the project. We acknowledge Shivangi Singhal (IIIT-Delhi, India) for providing us with the pre-processed multimodal fake news dataset. We also thank Bing He and Kartik Sharma for helping with translations, the CLAWS research group members for preliminary manual inspections of the translated text, and the anonymous reviewers for their constructive feedback.

\bibliography{bibfile}

\begin{thebibliography}{63}
\providecommand{\natexlab}[1]{#1}

\bibitem[{AAAS(2016)}]{american2016state}
AAAS. 2016.
\newblock The state of languages in the US: A statistical portrait.
\newblock
  \url{https://www.amacad.org/publication/state-languages-us-statistical-portrait}.
\newblock Accessed: 2022-01-09.

\bibitem[{Agarap(2018)}]{agarap2018deep}
Agarap, A.~F. 2018.
\newblock Deep learning using rectified linear units (ReLU).
\newblock \emph{arXiv:1803.08375}.

\bibitem[{Alam, Ofli, and Imran(2018)}]{alam2018crisismmd}
Alam, F.; Ofli, F.; and Imran, M. 2018.
\newblock Crisismmd: Multimodal twitter datasets from natural disasters.
\newblock In \emph{AAAI ICWSM}.

\bibitem[{Arviv, Hanouna, and Tsur(2021)}]{arviv2021sa}
Arviv, E.; Hanouna, S.; and Tsur, O. 2021.
\newblock It’sa Thin Line Between Love and Hate: Using the Echo in Modeling
  Dynamics of Racist Online Communities.
\newblock In \emph{AAAI ICWSM}.

\bibitem[{Basta, Costa-juss{\`a}, and Casas(2019)}]{basta2019evaluating}
Basta, C.; Costa-juss{\`a}, M.~R.; and Casas, N. 2019.
\newblock Evaluating the Underlying Gender Bias in Contextualized Word
  Embeddings.
\newblock In \emph{Proc. of the Workshop on Gender Bias in NLP}.

\bibitem[{Bender(2019)}]{bender2019rule}
Bender, E. 2019.
\newblock The \#BenderRule: On Naming the Languages We Study and Why It
  Matters.
\newblock \emph{The Gradient}.

\bibitem[{Bender et~al.(2021)Bender, Gebru, McMillan-Major, and
  Shmitchell}]{bender2021dangers}
Bender, E.~M.; Gebru, T.; McMillan-Major, A.; and Shmitchell, S. 2021.
\newblock On the Dangers of Stochastic Parrots: Can Language Models Be Too Big?
\newblock In \emph{ACM FAccT}.

\bibitem[{Caruana, Lawrence, and Giles(2000)}]{caruana2000overfitting}
Caruana, R.; Lawrence, S.; and Giles, C. 2000.
\newblock Overfitting in Neural Nets: Backpropagation, Conjugate Gradient, and
  Early Stopping.
\newblock \emph{NeurIPS}.

\bibitem[{Cañete et~al.(2020)Cañete, Chaperon, Fuentes, Ho, Kang, and
  Pérez}]{CaneteCFP2020}
Cañete, J.; Chaperon, G.; Fuentes, R.; Ho, J.-H.; Kang, H.; and Pérez, J.
  2020.
\newblock Spanish Pre-Trained BERT Model and Evaluation Data.
\newblock In \emph{PML4DC at ICLR 2020}.

\bibitem[{Choi et~al.(2021)Choi, Budak, Romero, and Jurgens}]{choi2021more}
Choi, M.; Budak, C.; Romero, D.~M.; and Jurgens, D. 2021.
\newblock More than Meets the Tie: Examining the Role of Interpersonal
  Relationships in Social Networks.
\newblock In \emph{AAAI ICWSM}.

\bibitem[{Conneau et~al.(2020)Conneau, Khandelwal, Goyal, Chaudhary, Wenzek,
  Guzm{\'a}n, Grave, Ott, Zettlemoyer, and Stoyanov}]{conneau2020unsupervised}
Conneau, A.; Khandelwal, K.; Goyal, N.; Chaudhary, V.; Wenzek, G.; Guzm{\'a}n,
  F.; Grave, {\'E}.; Ott, M.; Zettlemoyer, L.; and Stoyanov, V. 2020.
\newblock Unsupervised Cross-lingual Representation Learning at Scale.
\newblock In \emph{ACL}.

\bibitem[{Crook et~al.(2016)Crook, Glowacki, Suran, K.~Harris, and
  Bernhardt}]{crook2016content}
Crook, B.; Glowacki, E.~M.; Suran, M.; K.~Harris, J.; and Bernhardt, J.~M.
  2016.
\newblock Content analysis of a live CDC Twitter chat during the 2014 Ebola
  outbreak.
\newblock \emph{Comm'n. Res. Reports}.

\bibitem[{Cui et~al.(2020)Cui, Che, Liu, Qin, Wang, and
  Hu}]{cui-etal-2020-revisiting}
Cui, Y.; Che, W.; Liu, T.; Qin, B.; Wang, S.; and Hu, G. 2020.
\newblock Revisiting Pre-Trained Models for {C}hinese Natural Language
  Processing.
\newblock In \emph{EMNLP (Findings)}.

\bibitem[{De~Choudhury, Counts, and Horvitz(2013)}]{de2013social}
De~Choudhury, M.; Counts, S.; and Horvitz, E. 2013.
\newblock Social media as a measurement tool of depression in populations.
\newblock In \emph{ACM WebSci}.

\bibitem[{De~Choudhury et~al.(2013)De~Choudhury, Gamon, Counts, and
  Horvitz}]{de2013predicting}
De~Choudhury, M.; Gamon, M.; Counts, S.; and Horvitz, E. 2013.
\newblock Predicting depression via social media.
\newblock In \emph{AAAI ICWSM}.

\bibitem[{Deng et~al.(2009)Deng, Dong, Socher, Li, Li, and
  Fei-Fei}]{imagenet2009Deng}
Deng, J.; Dong, W.; Socher, R.; Li, L.-J.; Li, K.; and Fei-Fei, L. 2009.
\newblock ImageNet: A large-scale hierarchical image database.
\newblock In \emph{IEEE CVPR}.

\bibitem[{Devlin et~al.(2018)Devlin, Chang, Lee, and
  Toutanova}]{DBLP:journals/corr/abs-1810-04805}
Devlin, J.; Chang, M.; Lee, K.; and Toutanova, K. 2018.
\newblock {BERT:} Pre-training of Deep Bidirectional Transformers for Language
  Understanding.
\newblock \emph{CoRR}, abs/1810.04805.

\bibitem[{Doiron(2020)}]{HindiBERT}
Doiron, N. 2020.
\newblock {Hindi BERT on HuggingFace}.
\newblock \url{https://huggingface.co/monsoon-nlp/hindi-bert}.
\newblock Accessed: 2022-01-09.

\bibitem[{Duong, Lebret, and Aberer(2017)}]{duong2017multimodal}
Duong, C.~T.; Lebret, R.; and Aberer, K. 2017.
\newblock Multimodal classification for analysing social media.
\newblock \emph{arXiv:1708.02099}.

\bibitem[{Glasgow, Fink, and Boyd-Graber(2014)}]{glasgow2014our}
Glasgow, K.; Fink, C.; and Boyd-Graber, J. 2014.
\newblock " Our Grief is Unspeakable'': Automatically Measuring the Community
  Impact of a Tragedy.
\newblock In \emph{AAAI ICWSM}.

\bibitem[{Hedderich et~al.(2021)Hedderich, Lange, Adel, Str{\"o}tgen, and
  Klakow}]{hedderich2021survey}
Hedderich, M.~A.; Lange, L.; Adel, H.; Str{\"o}tgen, J.; and Klakow, D. 2021.
\newblock A Survey on Recent Approaches for Natural Language Processing in
  Low-Resource Scenarios.
\newblock In \emph{NAACL-HLT}.

\bibitem[{Hendricks et~al.(2018)Hendricks, Burns, Saenko, Darrell, and
  Rohrbach}]{hendricks2018women}
Hendricks, L.~A.; Burns, K.; Saenko, K.; Darrell, T.; and Rohrbach, A. 2018.
\newblock Women also snowboard: Overcoming bias in captioning models.
\newblock In \emph{Proceedings of the European Conference on Computer Vision
  (ECCV)}.

\bibitem[{Higgins et~al.(2009)Higgins, LaSalle, Zhaoxing, Kasten, Bing, Ridzon,
  and Witten}]{higgins2009validation}
Higgins, J.; LaSalle, A.; Zhaoxing, P.; Kasten, M.; Bing, K.; Ridzon, S.; and
  Witten, T. 2009.
\newblock Validation of photographic food records in children: are pictures
  really worth a thousand words?
\newblock \emph{Euro. J. of Clinical Nutrition}.

\bibitem[{Hong et~al.(2003)Hong, Benet-Martinez, Chiu, and
  Morris}]{hong2003boundaries}
Hong, Y.-Y.; Benet-Martinez, V.; Chiu, C.-Y.; and Morris, M.~W. 2003.
\newblock Boundaries of cultural influence: Construct activation as a mechanism
  for cultural differences in social perception.
\newblock \emph{J. of Cross-Cultural Psych.}

\bibitem[{Houston et~al.(2015)Houston, Hawthorne, Perreault, Park,
  Goldstein~Hode, Halliwell, Turner~McGowen, Davis, Vaid, McElderry
  et~al.}]{houston2015social}
Houston, J.~B.; Hawthorne, J.; Perreault, M.~F.; Park, E.~H.; Goldstein~Hode,
  M.; Halliwell, M.~R.; Turner~McGowen, S.~E.; Davis, R.; Vaid, S.; McElderry,
  J.~A.; et~al. 2015.
\newblock Social media and disasters: a functional framework for social media
  use in disaster planning, response, and research.
\newblock \emph{Disasters}.

\bibitem[{Islam and Goldwasser(2021)}]{islam2021analysis}
Islam, T.; and Goldwasser, D. 2021.
\newblock Analysis of Twitter Users' Lifestyle Choices using Joint Embedding
  Model.
\newblock In \emph{AAAI ICWSM}.

\bibitem[{Joshi et~al.(2020)Joshi, Santy, Budhiraja, Bali, and
  Choudhury}]{joshi2020state}
Joshi, P.; Santy, S.; Budhiraja, A.; Bali, K.; and Choudhury, M. 2020.
\newblock The State and Fate of Linguistic Diversity and Inclusion in the NLP
  World.
\newblock In \emph{ACL}.

\bibitem[{Junczys-Dowmunt et~al.(2018)Junczys-Dowmunt, Grundkiewicz, Dwojak,
  Hoang, Heafield, Neckermann, Seide, Germann, Fikri~Aji, Bogoychev, Martins,
  and Birch}]{mariannmt}
Junczys-Dowmunt, M.; Grundkiewicz, R.; Dwojak, T.; Hoang, H.; Heafield, K.;
  Neckermann, T.; Seide, F.; Germann, U.; Fikri~Aji, A.; Bogoychev, N.;
  Martins, A. F.~T.; and Birch, A. 2018.
\newblock Marian: Fast Neural Machine Translation in {C++}.
\newblock In \emph{ACL 2018, System Demonstrations}.

\bibitem[{Kern et~al.(2016)Kern, Park, Eichstaedt, Schwartz, Sap, Smith, and
  Ungar}]{kern2016gaining}
Kern, M.~L.; Park, G.; Eichstaedt, J.~C.; Schwartz, H.~A.; Sap, M.; Smith,
  L.~K.; and Ungar, L.~H. 2016.
\newblock Gaining insights from social media language: Methodologies and
  challenges.
\newblock \emph{Psych. Methods}.

\bibitem[{Kingma and Ba(2014)}]{kingma2014adam}
Kingma, D.~P.; and Ba, J. 2014.
\newblock Adam: A method for stochastic optimization.
\newblock \emph{arXiv:1412.6980}.

\bibitem[{Lazer et~al.(2018)Lazer, Baum, Benkler, Berinsky, Greenhill, Menczer,
  Metzger, Nyhan, Pennycook, Rothschild et~al.}]{lazer2018science}
Lazer, D.~M.; Baum, M.~A.; Benkler, Y.; Berinsky, A.~J.; Greenhill, K.~M.;
  Menczer, F.; Metzger, M.~J.; Nyhan, B.; Pennycook, G.; Rothschild, D.; et~al.
  2018.
\newblock The science of fake news.
\newblock \emph{Science}.

\bibitem[{Li et~al.(2020)Li, Peng, Li, Xia, Yang, Sun, Yu, and
  He}]{li2020survey}
Li, Q.; Peng, H.; Li, J.; Xia, C.; Yang, R.; Sun, L.; Yu, P.~S.; and He, L.
  2020.
\newblock A survey on text classification: From shallow to deep learning.
\newblock \emph{arXiv:2008.00364}.

\bibitem[{Liu et~al.(2021)Liu, Bugliarello, Ponti, Reddy, Collier, and
  Elliott}]{liu2021visually}
Liu, F.; Bugliarello, E.; Ponti, E.~M.; Reddy, S.; Collier, N.; and Elliott, D.
  2021.
\newblock Visually Grounded Reasoning across Languages and Cultures.
\newblock In \emph{EMNLP}.

\bibitem[{MacLaughlin et~al.(2021)MacLaughlin, Chen, Ayan, and
  Roth}]{maclaughlin2021context}
MacLaughlin, A.; Chen, T.; Ayan, B.~K.; and Roth, D. 2021.
\newblock Context-based quotation recommendation.
\newblock In \emph{AAAI}.

\bibitem[{Martin et~al.(2020)Martin, Muller, Ortiz~Su{\'a}rez, Dupont, Romary,
  de~la Clergerie, Seddah, and Sagot}]{martin-etal-2020-camembert}
Martin, L.; Muller, B.; Ortiz~Su{\'a}rez, P.~J.; Dupont, Y.; Romary, L.; de~la
  Clergerie, {\'E}.; Seddah, D.; and Sagot, B. 2020.
\newblock {C}amem{BERT}: a Tasty {F}rench Language Model.
\newblock In \emph{ACL}.

\bibitem[{Microsoft(2019)}]{microsoft}
Microsoft. 2019.
\newblock {Neural Machine Translation Enabling Human Parity Innovations In the
  Cloud}.
\newblock
  \url{https://www.microsoft.com/en-us/translator/blog/2019/06/17/neural-machine-translation-enabling-human-parity-innovations-in-the-cloud/}.
\newblock Accessed: 2022-01-09.

\bibitem[{Mielke(2016)}]{sabrina2016language}
Mielke, S.~J. 2016.
\newblock Language diversity in {ACL} 2004 - 2016.
\newblock \url{https://sjmielke.com/acl-language-diversity.htm}.
\newblock Accessed: 2022-01-09.

\bibitem[{Mocanu et~al.(2013)Mocanu, Baronchelli, Perra, Gon{\c{c}}alves,
  Zhang, and Vespignani}]{mocanu2013twitter}
Mocanu, D.; Baronchelli, A.; Perra, N.; Gon{\c{c}}alves, B.; Zhang, Q.; and
  Vespignani, A. 2013.
\newblock The twitter of babel: Mapping world languages through microblogging
  platforms.
\newblock \emph{PloS One}.

\bibitem[{Muller et~al.(2021)Muller, Anastasopoulos, Sagot, and
  Seddah}]{muller2021being}
Muller, B.; Anastasopoulos, A.; Sagot, B.; and Seddah, D. 2021.
\newblock When Being Unseen from mBERT is just the Beginning: Handling New
  Languages With Multilingual Language Models.
\newblock In \emph{NAACL-HLT}.

\bibitem[{Ngiam et~al.(2011)Ngiam, Khosla, Kim, Nam, Lee, and
  Ng}]{ngiam2011multimodal}
Ngiam, J.; Khosla, A.; Kim, M.; Nam, J.; Lee, H.; and Ng, A.~Y. 2011.
\newblock Multimodal deep learning.
\newblock In \emph{ICML}.

\bibitem[{Nozza, Bianchi, and Hovy(2020)}]{nozza2020mask}
Nozza, D.; Bianchi, F.; and Hovy, D. 2020.
\newblock What the [mask]? making sense of language-specific BERT models.
\newblock \emph{arXiv:2003.02912}.

\bibitem[{Ofli, Alam, and Imran(2020)}]{ofli2020analysis}
Ofli, F.; Alam, F.; and Imran, M. 2020.
\newblock Analysis of Social Media Data using Multimodal Deep Learning for
  Disaster Response.
\newblock \emph{arXiv:2004.11838}.

\bibitem[{Pennebaker, Francis, and Booth(2001)}]{pennebaker2001linguistic}
Pennebaker, J.~W.; Francis, M.~E.; and Booth, R.~J. 2001.
\newblock Linguistic inquiry and word count: LIWC 2001.
\newblock \emph{Mahway: Lawrence Erlbaum Associates}.

\bibitem[{PewResearch(2018)}]{pew_latinos}
PewResearch. 2018.
\newblock Latinos and discrimination.
\newblock
  \url{https://www.pewresearch.org/hispanic/2018/10/25/latinos-and-discrimination/}.
\newblock Accessed: 2021-09-10.

\bibitem[{Pires, Schlinger, and Garrette(2019)}]{pires2019multilingual}
Pires, T.; Schlinger, E.; and Garrette, D. 2019.
\newblock How Multilingual is Multilingual BERT?
\newblock In \emph{ACL}.

\bibitem[{Raffel et~al.(2020)Raffel, Shazeer, Roberts, Lee, Narang, Matena,
  Zhou, Li, and Liu}]{raffel2020exploring}
Raffel, C.; Shazeer, N.; Roberts, A.; Lee, K.; Narang, S.; Matena, M.; Zhou,
  Y.; Li, W.; and Liu, P.~J. 2020.
\newblock Exploring the Limits of Transfer Learning with a Unified Text-to-Text
  Transformer.
\newblock \emph{JMLR}.

\bibitem[{Rasmy et~al.(2021)Rasmy, Xiang, Xie, Tao, and Zhi}]{rasmy2021med}
Rasmy, L.; Xiang, Y.; Xie, Z.; Tao, C.; and Zhi, D. 2021.
\newblock Med-BERT: pretrained contextualized embeddings on large-scale
  structured electronic health records for disease prediction.
\newblock \emph{NPJ Digital Med.}

\bibitem[{Reynolds and Johnson(2011)}]{reynolds2011picture}
Reynolds, L.; and Johnson, R. 2011.
\newblock Is a picture is worth a thousand words? Creating effective
  questionnaires with pictures.
\newblock \emph{Practical Assessment, Research, and Eval.}

\bibitem[{Sanh et~al.(2019)Sanh, Debut, Chaumond, and
  Wolf}]{sanh2019distilbert}
Sanh, V.; Debut, L.; Chaumond, J.; and Wolf, T. 2019.
\newblock DistilBERT, a distilled version of BERT: smaller, faster, cheaper and
  lighter.
\newblock \emph{arXiv:1910.01108}.

\bibitem[{Sellam et~al.(2021)Sellam, Yadlowsky, Wei, Saphra, D'Amour, Linzen,
  Bastings, Turc, Eisenstein, Das et~al.}]{sellam2021multiberts}
Sellam, T.; Yadlowsky, S.; Wei, J.; Saphra, N.; D'Amour, A.; Linzen, T.;
  Bastings, J.; Turc, I.; Eisenstein, J.; Das, D.; et~al. 2021.
\newblock The MultiBERTs: BERT Reproductions for Robustness Analysis.
\newblock \emph{arXiv:2106.16163}.

\bibitem[{Shen, Wilson, and Mihalcea(2019)}]{shen2019measuring}
Shen, Y.; Wilson, S.~R.; and Mihalcea, R. 2019.
\newblock Measuring personal values in cross-cultural user-generated content.
\newblock In \emph{Int. Conf. on Social Informatics}. Springer.

\bibitem[{Shu et~al.(2018)Shu, Mahudeswaran, Wang, Lee, and
  Liu}]{shu2018fakenewsnet}
Shu, K.; Mahudeswaran, D.; Wang, S.; Lee, D.; and Liu, H. 2018.
\newblock FakeNewsNet: A Data Repository with News Content, Social Context and
  Dynamic Information for Studying Fake News on Social Media.
\newblock \emph{arXiv:1809.01286}.

\bibitem[{Shu et~al.(2017)Shu, Sliva, Wang, Tang, and Liu}]{shu2017fake}
Shu, K.; Sliva, A.; Wang, S.; Tang, J.; and Liu, H. 2017.
\newblock Fake News Detection on Social Media: A Data Mining Perspective.
\newblock \emph{ACM SIGKDD Explorations Newsletter}.

\bibitem[{Simonyan and Zisserman(2015)}]{simonyanVeryDeepConvolutional2015a}
Simonyan, K.; and Zisserman, A. 2015.
\newblock Very {{Deep Convolutional Networks}} for {{Large}}-{{Scale Image
  Recognition}}.
\newblock \emph{arXiv:1409.1556}.

\bibitem[{Singhal et~al.(2020)Singhal, Kabra, Sharma, Shah, Chakraborty, and
  Kumaraguru}]{singhal2020spotfake+}
Singhal, S.; Kabra, A.; Sharma, M.; Shah, R.~R.; Chakraborty, T.; and
  Kumaraguru, P. 2020.
\newblock Spotfake+: A multimodal framework for fake news detection via
  transfer learning (student abstract).
\newblock In \emph{AAAI}.

\bibitem[{Singhal et~al.(2019)Singhal, Shah, Chakraborty, Kumaraguru, and
  Satoh}]{singhal2019spotfake}
Singhal, S.; Shah, R.~R.; Chakraborty, T.; Kumaraguru, P.; and Satoh, S. 2019.
\newblock Spotfake: A multi-modal framework for fake news detection.
\newblock In \emph{IEEE Int. Cont. on Multimedia Big Data (BigMM)}. IEEE.

\bibitem[{Souza, Nogueira, and Lotufo(2020)}]{souza2020bertimbau}
Souza, F.; Nogueira, R.; and Lotufo, R. 2020.
\newblock {BERT}imbau: pretrained {BERT} models for {B}razilian {P}ortuguese.
\newblock In \emph{9th Brazilian Conference on Intelligent Systems {BRACIS}}.

\bibitem[{Strubell, Ganesh, and McCallum(2019)}]{strubell2019energy}
Strubell, E.; Ganesh, A.; and McCallum, A. 2019.
\newblock Energy and Policy Considerations for Deep Learning in NLP.
\newblock In \emph{ACL}.

\bibitem[{Sun, Huang, and Qiu(2019)}]{sun2019utilizing}
Sun, C.; Huang, L.; and Qiu, X. 2019.
\newblock Utilizing BERT for Aspect-Based Sentiment Analysis via Constructing
  Auxiliary Sentence.
\newblock In \emph{NAACL-HLT}.

\bibitem[{Teixeira, Wedel, and Pieters(2012)}]{teixeira2012emotion}
Teixeira, T.; Wedel, M.; and Pieters, R. 2012.
\newblock Emotion-induced engagement in internet video advertisements.
\newblock \emph{J. of Marketing Res.}

\bibitem[{Wolf et~al.(2019)Wolf, Debut, Sanh, Chaumond, Delangue, Moi, Cistac,
  Rault, Louf, Funtowicz et~al.}]{wolf2019huggingface}
Wolf, T.; Debut, L.; Sanh, V.; Chaumond, J.; Delangue, C.; Moi, A.; Cistac, P.;
  Rault, T.; Louf, R.; Funtowicz, M.; et~al. 2019.
\newblock Huggingface's transformers: State-of-the-art natural language
  processing.
\newblock \emph{arXiv:1910.03771}.

\bibitem[{Wu and Dredze(2020)}]{wu2020all}
Wu, S.; and Dredze, M. 2020.
\newblock Are All Languages Created Equal in Multilingual BERT?
\newblock In \emph{Proc. of the Workshop on Representation Learning for NLP}.

\bibitem[{Xue et~al.(2021)Xue, Constant, Roberts, Kale, Al-Rfou, Siddhant,
  Barua, and Raffel}]{xue2021mt5}
Xue, L.; Constant, N.; Roberts, A.; Kale, M.; Al-Rfou, R.; Siddhant, A.; Barua,
  A.; and Raffel, C. 2021.
\newblock mT5: A Massively Multilingual Pre-trained Text-to-Text Transformer.
\newblock In \emph{NAACL-HLT}.

\end{thebibliography}

\end{document}